\documentclass[a4 paper, 12pt]{article}

\usepackage[margin=1in,footskip=0.5in]{geometry}
\usepackage{microtype}
\usepackage{graphicx}
\usepackage{subfigure}
\usepackage{booktabs} 
\usepackage{tabularx}

\usepackage{authblk}
\usepackage{hyperref}
\usepackage{makecell}
\usepackage{arydshln}
\usepackage{array}

\usepackage{multirow}  
\usepackage{tikz}      
\usepackage{comment}
\usepackage[numbers]{natbib}

\usepackage{amsmath}
\usepackage{amssymb}
\usepackage{mathtools}
\usepackage{amsthm}

\usepackage[capitalize,noabbrev]{cleveref}

\usepackage{enumitem}
\usepackage{bbm}

\usepackage{xcolor}
\usepackage{pifont}
\definecolor{myred}{RGB}{215,48,39}
\definecolor{mygreen}{RGB}{26,152,80}
\newcommand{\cmark}{\textcolor{mygreen}{\ding{51}}}
\newcommand{\xmark}{\textcolor{myred}{\ding{55}}}

\newtheorem{theorem}{Theorem}
\newtheorem{lemma}{Lemma}
\newtheorem{proposition}{Proposition}
\newtheorem{definition}{Definition}

\newtheorem{remark}{Remark}

\usepackage[textsize=tiny]{todonotes}

\allowdisplaybreaks[4]

\title{Transformers Can Overcome the Curse of Dimensionality: 
A Theoretical Study from an Approximation Perspective}

\author[a,b]{Yuling Jiao}
\author[c]{Yanming Lai\thanks{Corresponding Author (ylaiam@connect.ust.hk)}}
\author[c]{Yang Wang}
\author[c]{Bokai Yan}
\affil[a]{School of Artificial Intelligence, Wuhan University, Wuhan, 430072, Hubei Province, China}
\affil[b]{Hubei Key Laboratory of Computational Science, Wuhan University, Wuhan, 430072, Hubei Province, China}
\affil[c]{Department of Mathematics, The Hong Kong University of Science and Technology, Clear Water Bay, Kowloon, Hong Kong, China }

\date{}

\begin{document}

\maketitle


\begin{abstract}

The Transformer model is widely used in various application areas of machine learning, such as natural language processing. This paper investigates the approximation of the Hölder continuous function class $\mathcal{H}_{Q}^{\beta}\left([0,1]^{d\times n},\mathbb{R}^{d\times n}\right)$ by Transformers and constructs several Transformers that can overcome the curse of dimensionality. These Transformers consist of one self-attention layer with one head and the softmax function as the activation function, along with several feedforward layers. For example, to achieve an approximation accuracy of $\epsilon$, if the activation functions of the feedforward layers in the Transformer are ReLU and floor, only $\mathcal{O}\left(\log\frac{1}{\epsilon}\right)$ layers of feedforward layers are needed, with widths of these layers not exceeding $\mathcal{O}\left(\frac{1}{\epsilon^{2/\beta}}\log\frac{1}{\epsilon}\right)$. If other activation functions are allowed in the feedforward layers, the width of the feedforward layers can be further reduced to a constant. These results demonstrate that Transformers have a strong expressive capability. The construction in this paper is based on the Kolmogorov-Arnold Representation Theorem and does not require the concept of contextual mapping, hence our proof is more intuitively clear compared to previous Transformer approximation works. Additionally, the translation technique proposed in this paper helps to apply the previous approximation results of feedforward neural networks to Transformer research.

\end{abstract}

\section{Introduction}

The Transformer model with the self-attention mechanism, proposed by \cite{vaswani2017attention}, has been widely applied in deep learning since its inception. Its extensive applications span across multiple fields, including natural language processing \cite{radford2018improving,radford2019language,devlin2018bert,yang2020xlnetgeneralizedautoregressivepretraining,liu2019robertarobustlyoptimizedbert,brown2020language,fedus2022switch}, image processing \cite{dosovitskiy2021an,ying2021transformers}, generative tasks \cite{peebles2023scalable}, reinforcement learning \cite{parisotto2020stabilizing,chen2021decision,janner2021offline}, video understanding \cite{akbari2021vatt,DBLP:conf/icml/BertasiusWT21}, and more. The main advantage of Transformers lies in their efficient scalability, which is achieved through parallel token processing, parameter sharing, and token interaction based on simple dot products. Despite its successes in various domains, a comprehensive theoretical understanding of the Transformer model is still evolving.

One of the key reasons for the success of neural network models, including Transformer models, is their ability to broadly represent various functions. The universal approximation theorem is a classic result in neural network theory that dates back several decades \cite{cybenko1989approximation,hornik1991approximation,funahashi1989approximate}. These results indicate that as long as the width is large enough, a two-layer feedforward neural network (FNN) can approximate any continuous function with compact support to arbitrary precision. 
In recent years, there has been significant progress in the theoretical study of approximation for FNNs. From shallow to deep networks, from ReLU to sigmoid activations, various studies have explored the approximation capabilities of different FNNs. The readers are referred to \cite{yarotsky2017error,lu2017expressive,yarotsky2018optimal,DBLP:conf/iclr/Suzuki19,shen2020deep,lu2021deep,shen2022optimal,guhring2020error,guhring2021approximation,siegel2024sharp} and the references therein. These studies cover a variety of function spaces such as Holder continuous spaces, \(C^s\) spaces, Sobolev spaces, Besov spaces, Barron spaces, revealing the broad expressive power of FNNs. Furthermore, some research indicates that by allowing activations other than ReLU and sigmoid in network settings, constructed FNNs can overcome the curse of dimensionality, which is a term proposed by Bellman \cite{bellman1952theory} to depict an exponentially increasing difficulty of problems with increasing dimension. These FNNs include such as ReLU-floor networks \cite{shen2021deep}, ReLU-\(2^x\)-sine networks \cite{jiao2023deep}, ReLU-sine networks \cite{yarotsky2020phase}, two types of three-layer networks proposed by \cite{shen2021neural}, where the sizes of all these networks do not grow exponentially with the dimensionality. Interestingly, \cite{yarotsky2021elementary,zhang2022deep,wang2024don} have even constructed several networks with their size independent of the approximation accuracy.

Studying the expressive power of Transformers is more challenging compared to FNNs. The main challenge in proving the universal approximation theorem for Transformer models is that Transformers need to consider the context of the entire input sequence. Unlike in FNNs where each input is processed independently, the self-attention mechanism in Transformer models must consider the dependencies between all tokens in each input sequence. \cite{DBLP:conf/iclr/YunBRRK20} introduced the concept of “contextual mapping" to aggregate these dependencies into a token-based quantity through the self-attention mechanism, which is then mapped by a feedforward neural network to the desired output, thus first proving the universal approximation theorem for Transformer models. They demonstrated that if the number of self-attention layers in a Transformer is proportional to the power of the length $n$ of each input sequence, it can approximate continuous functions on compact support regions. This result was later extended to sparse-attention Transformers \cite{zaheer2020big,yun2020n} and Transformers with constraints \cite{DBLP:conf/iclr/KratsiosZLD22}. \cite{kim2023provable} improved the results of \cite{DBLP:conf/iclr/YunBRRK20}, showing that $2n$ layers of self-attention are sufficient for memorizing finite samples. Based on the concept of contextual mapping and the properties of the Boltzmann operator composed of softmax, \cite{kajitsukatransformers} demonstrated that a single-layer Transformer with only one head has memorization capabilities, and a Transformer composed of one self-attention layer and two feedforward layers can approximate permutation equivariant functions on compact sets. {\cite{hu2024fundamental} extended the results of \cite{kajitsukatransformers} by generalizing the latter's rank-1 weight matrices to weight matrices of arbitrary rank.} Recently, \cite{jiang2025approximationratetransformerarchitecture} explored the approximation capabilities of Transformer networks for Hölder and Sobolev functions, and apply these results to address nonparametric regression estimation with dependent observations.

In addition, \cite{wei2022statistically} investigated the expressive power of Transformers under statistical learnability constraints. \cite{gurevych2022rate} analyzed the theoretical performance of Transformers from the perspective of a hierarchical compositional model. Subsequently, \cite{takakura2023approximation} extended their results, demonstrating that a one-layer Transformer with an embedding layer is the universal approximator of a shift-equivariant $\gamma$ smooth function. \cite{petrov2024prompting} showed that prompting and prefix-tuning a pretrained model can universally approximate sequence-to-sequence functions. \cite{luo2022your} proved that there exist functions that cannot be approximated by Transformers with positional encodings. From the perspective of sequence modeling, \cite{wang2024understandingexpressivepowermechanisms,jiang2025approximationratetransformerarchitecture} investigated the expressive power of the Transformer and compared it with that of RNNs. Works on Transformer approximation of sparse functions include \cite{edelman2022inductive,DBLP:conf/acl/BhattamishraPKB23,sanford2024representational,trauger2024sequence,wang2024transformers}. More research on the memory capabilities of Transformers can be found in \cite{DBLP:conf/iclr/MahdaviLT24,madden2024upper,DBLP:conf/icml/ChenZ24a,kajitsuka2024optimal}.

There are some other works that studied the expressive power of Transformers from different perspectives. \cite{vuckovic2020mathematical} and \cite{kim2021lipschitz} investigated the Lipschitz smoothness of attention operations. \cite{bhojanapalli2020low} demonstrated that the small size of attention heads limits the rank of the self-attention matrix. \cite{wies2021transformer} pinpointed the rank of the input embedding matrix as a bottleneck limiting the impact of network width on expressive capacity. \cite{dong2021attention} showed that without skip connections or feedforward layers, the rank of self-attention layers decreases exponentially. \cite{likhosherstov2023expressive} highlighted that self-attention modules have the ability to replicate any sparse pattern.

\begin{table*}[t]
\centering
\resizebox{\textwidth}{!}{
\begin{tabular}{c | c c c}
\multicolumn{1}{c|}{\textbf{Reference}} & \textbf{Use Softmax?} & \textbf{Approximation Rate?} & \makecell[c]{\textbf{Overcome the} \\ \textbf{Curse of} \\ \textbf{Dimensionality?}} \\
\hline
\citet{DBLP:conf/iclr/YunBRRK20}       & \cmark & \xmark & \xmark \\
\citet{kajitsukatransformers}          & \cmark & \xmark & \xmark \\
\citet{fang2022attention}             & \xmark & \xmark & \xmark \\
\citet{gurevych2022rate}               & \xmark & \cmark & \cmark \\
\citet{jiao2024convergence}            & \xmark & \cmark & \xmark \\
\citet{havrilla2024understanding}      & \xmark & \cmark & \xmark \\
\citet{jiaolaisunwangyan}              & \cmark & \cmark & \xmark \\
\textbf{Ours} (Theorem \ref{Linfty}) & \cmark & \cmark & \cmark \\
\textbf{Ours} (Theorem \ref{Lp})    & \cmark & \cmark & \cmark \\
\end{tabular}\vspace{-5pt}
}
\caption{Comparison of approximation capabilities among various Transformer models.}
\label{table: 1}
\end{table*}

\subsection{Our Contributions}

The seminal works on transformer approximation \cite{DBLP:conf/iclr/YunBRRK20,kim2023provable,kajitsukatransformers} focused on the approximation of continuous functions using Transformers, with results all suffering from the curse of dimensionality. A natural question arises: can we construct a Transformer that overcomes the curse of dimensionality under the weakest possible assumptions about the target function? Can the previously established FNNs that have overcome the curse of dimensionality be utilized by us? This paper addresses these questions. Starting from the Kolmogorov–Arnold Representation Theorem, this paper transforms the approximation problem of Transformers into approximation and memorization problems of FNNs, constructing several Transformers that can overcome the curse of dimensionality. This reveals the exceptional expressive power of Transformers and partly explains the significant success of Transformers in practical applications. Specifically, the contributions of this paper are as follows:
\begin{itemize}
\item This paper constructs several transformers that overcome the curse of dimensionality. To the best of the author's knowledge, this is the first work demonstrating that transformers can overcome the curse of dimensionality. Furthermore, based on the construction presented in this paper, more transformers of this kind can be generated. The constructed transformers require only one layer of self-attention, with both the depth and width of the feedforward layers being small. We only need to assume that the target function is Hölder continuous, which is a very weak assumption. Therefore, the results of this paper have a broad range of applicability.

\item Previous works solely focused on approximations within the \(L^{p}\) distance. This paper not only introduces transformers capable of approximating the target function within the \(L^{p}\) distance (Theorem \ref{Lp}), but also develops transformers that can approximate the target function within the \(L^{\infty}\) distance (Theorem \ref{Linfty}).

\item The construction in this paper is based on the Kolmogorov–Arnold Representation Theorem and does not require the use of the concept of contextual mapping. Therefore, the proof process in this paper is more intuitive and clear compared to previous transformer approximation works. This offers a new approach to studying the approximation properties of transformers.

\item How to directly apply the previous results of FNNs to the study of transformers is an important technical issue addressed in this paper. In the conventional setting of the feedforward layer in transformers, the layer applies the same operation to each column of the input matrix. This paper introduces a translation technique that allows the feedforward layer of the transformer to perform different operations on each column of the input matrix, enabling the application of the Kolmogorov–Arnold Representation Theorem. For details, please refer to Section \ref{basic idea}.
\end{itemize}

It is worth noting that \cite{jiang2025approximationratetransformerarchitecture,petrov2024prompting} also employed the Kolmogorov–Arnold Representation Theorem to investigate the expressive power of Transformers. However, there are several fundamental differences between our work and theirs. The differences between us and \cite{jiang2025approximationratetransformerarchitecture} are as follows. First, the problem formulations differ: we study an approximation problem mapping from $\mathbb{R}^{d\times n}$ to $\mathbb{R}^{d\times n}$, whereas they examine sequence modeling problems involving temporal indices $s$; Second, the Transformer architectures differ: our definition aligns with classical theoretical works \cite{DBLP:conf/iclr/YunBRRK20,kajitsukatransformers}, while they adopt alternative formulations that notably omit specification of activation functions in feedforward layers. The differences between us and \cite{petrov2024prompting} are as follows. First, their required parameter complexity is $\mathcal{O}(\epsilon^{-10-14m-4m^2})$, with $m$ being the spatial dimension, whereas our maximum parameter complexity is $\mathcal{O}\left(\left(\frac{1}{\epsilon}\right)^{4/\beta}\log^3\frac{1}{\epsilon}\right)$. Therefore, our results successfully overcome the curse of dimensionality while theirs do not. Second, our application of KST differs from theirs, and our proof techniques are entirely independent. We employ translation techniques, while they utilize what they term ``core attention head" methodology. Third, their required number of attention layers exhibits linear dependence on input sequence length, while our approach requires only a single attention layer. Given these fundamental differences, our work constitutes an independent contribution.

We compare our work with existing studies in Table \ref{table: 1}. A more detailed comparison and a summary of our main results are provided in Table \ref{comp}.

\subsection{Organization of This Paper}

The organization of this paper is as follows. In Section \ref{reliminaries}, we introduce the architecture of our transformer, the Kolmogorov-Arnold Representation Theorem, and the notations we use. Our main results is presented in Section \ref{main results}, along with the basic ideas of our construction and formal proofs. In Section \ref{Conclusions}, we summarize and conclude the paper. The proofs of all auxiliary lemmas can be found in the appendix.

\section{Preliminaries}\label{reliminaries}

\subsection{Transformer}


The Transformer model, first introduced by \cite{vaswani2017attention}, is a mapping which takes a sequence $\boldsymbol{X} \in \mathbb{R}^{d \times n}$ composed of $n$ tokens each with an embedding dimension of size $d$ as an input and outputs another sequence with the same size. In a Transformer model, there are two primary layers: a self-attention layer and a token-wise feedforward layer. The self-attention layer recalibrates each token embedding by aggregating information from all tokens in the sequence, calculated through dot-products of token embeddings. Subsequently, the token-wise feed-forward layer operates independently on these adjusted token embeddings without inter-token communication. Notably, Transformers utilize pairwise dot-products exclusively to capture token interactions within the input sequence.

The Transformer we focus on in this paper is composed of alternating feedforward blocks and self-attention layers. A self-attention layer $\boldsymbol{\mathcal{F}}_{SA}:\mathbb{R}^{d\times n}\to\mathbb{R}^{d\times n}$ with $h$ heads is defined as
\begin{align*}
&\boldsymbol{\mathcal{F}}_{SA}(\boldsymbol{X}):=
\boldsymbol{X}+\sum_{i=1}^h \boldsymbol{W}_O^{(i)}\boldsymbol{W}_V^{(i)} \boldsymbol{X} \sigma_S\left[\left(\boldsymbol{W}_K^{(i)} \boldsymbol{X}\right)^{\top}\left(\boldsymbol{W}_Q^{(i)} \boldsymbol{X}\right)\right],
\end{align*}
where $\boldsymbol{W}_V^{(i)}, \boldsymbol{W}_K^{(i)}, \boldsymbol{W}_Q^{(i)} \in \mathbb{R}^{s \times d}$ and $\boldsymbol{W}_O^{(i)} \in \mathbb{R}^{d \times s}$ are the weight matrices, $s$ is the head size, and $\sigma_S:\mathbb{R}^{d}\to\mathbb{R}^{d}$ is the softmax operator with output $[\sigma_S(\boldsymbol{x})]_i={e^{x_i}}/{\sum_{i=1}^d e^{x_i}},i\in[d]$ when acting on a vector $\boldsymbol{x}$. Here $\sigma_S$ acts on the input matrix column-wise. For the sake of simplicity, similar to \cite{DBLP:conf/iclr/YunBRRK20,kim2023provable,kajitsukatransformers}, we excludes the layer normalization from the original definition of \cite{vaswani2017attention} here.

A feedforward block $\boldsymbol{\mathcal{F}}_{FF}:\mathbb{R}^{d\times n}\to\mathbb{R}^{d\times n}$ with depth $L$ and element-wise activation $\sigma$ is defined by
\begin{align*}
\boldsymbol{\mathcal{F}}_0&=\boldsymbol{X};\\
\boldsymbol{\mathcal{F}}_{l}&=\sigma(\boldsymbol{W}_{l}\boldsymbol{\mathcal{F}}_{l-1}+\boldsymbol{B}_l),\quad l\in\{1,2,\dots,L-1\};\\
\boldsymbol{\mathcal{F}}_{FF}&=\boldsymbol{W}_{L}\boldsymbol{\mathcal{F}}_{L-1}.
\end{align*}
where $\boldsymbol{W}_{l} \in \mathbb{R}^{n_l \times n_{l-1}}$ are the weight matrices, and $\boldsymbol{B}_{l} \in \mathbb{R}^{n_l \times n}$ are the bias matrices, $n_l$ are the hidden dimensions. It can be observed that the feedforward block we define does not include skip connections. In comparison to the standard definition, we also allow the bias matrices $\boldsymbol{B}_l$ to have different columns. In the case where the columns of $\boldsymbol{B}_l$ are identical vectors $\boldsymbol{b}_l$: $\boldsymbol{B}_l=\boldsymbol{b}_l\boldsymbol{1}_{1\times n}$, we say that $\boldsymbol{\mathcal{F}}_{FF}$ is generated from a feedforward block $\boldsymbol{f}_{FF}$ acting on vectors, where \(\boldsymbol{f}_{FF}:\mathbb{R}^{d}\to\mathbb{R}^{d}\) is defined in the following way:
\begin{align*}
\boldsymbol{f}_0&=\boldsymbol{x};\\
\boldsymbol{f}_{l}&=\sigma(\boldsymbol{W}_{l}\boldsymbol{f}_{l-1}+\boldsymbol{b}_l),\quad l\in\{1,2,\dots,L-1\};\\
\boldsymbol{f}_{FF}&=\boldsymbol{W}_{L}\boldsymbol{f}_{L-1}.
\end{align*}

With these definitions, the Transformer $\boldsymbol{T}:\mathbb{R}^{d\times n}\to\mathbb{R}^{d\times n}$ is composed of alternating feedforward blocks and self-attention layers, with feedforward blocks at both the beginning and the end:
\begin{align*}
\boldsymbol{T}:=
\boldsymbol{\mathcal{F}}_{FF}^{(N)}\circ\boldsymbol{\mathcal{F}}_{SA}^{(N)}\circ\boldsymbol{\mathcal{F}}_{FF}^{(N-1)}\circ\cdots\circ\boldsymbol{\mathcal{F}}_{FF}^{(1)}\circ\boldsymbol{\mathcal{F}}_{SA}^{(1)}\circ\boldsymbol{\mathcal{F}}_{FF}^{(0)}.
\end{align*}

\subsection{Kolmogorov–Arnold Representation Theorem}

Kolmogorov-Arnold Representation Theorem (KST), first proposed by Kolmogorov in 1957 \cite{kolmogorov1957representation}, states that any continuous multivariate function can be represented as a superposition of continuous functions of fewer variables. This theorem provides a fundamental insight into the representation of functions in high-dimensional spaces, offering a way to simplify complex functions by breaking them down into simpler components. It is worth mentioning that KST is closely related to Hilbert's Thirteenth Problem.

Below we present the original KST:
\begin{proposition}[KST]
There exist univariate continuous functions $\psi_{p, q}$ such that for any continuous function $f:[0,1]^d\to\mathbb{R}$, there exist univariate continuous functions $g_q$ such that
\begin{align*}
f\left(x_1, \ldots, x_d\right)=\sum_{q=0}^{2 d} g_q\left(\sum_{p=1}^d \psi_{p, q}\left(x_p\right)\right).
\end{align*}
\end{proposition}

Since the proposal of KST, a great number of variants have emerged. In 2021, Hieber established a type of KST using the technique of space-filling curves, which can transfer the smoothness properties of the target function to the outer function \cite{schmidt2021kolmogorov}. This discovery enables researchers to apply KST to investigate the approximation properties of FNNs. Our work is also built upon his result.
\begin{definition}[Hölder continuous function]\label{holder}
Let $\beta \leq 1,Q\in\mathbb{R}$. $f:[0,1]^{d}\to \mathbb{R}$ is called $(\beta,Q)$-Hölder continuous function if for any $\boldsymbol{x},\boldsymbol{y}\in [0,1]^{d}$, there holds
\begin{align}\label{holder1}
|f(\boldsymbol{x})-f(\boldsymbol{y})|\leq Q\|\boldsymbol{x}-\boldsymbol{y}\|_{\infty}^\beta.
\end{align}
where $\|\cdot\|_{\infty}$ is the infty norm of vectors. $\mathcal{H}_{Q}^{\beta}\left([0,1]^{d},\mathbb{R}\right)$ denotes the class that consisting all the $(\beta,Q)$-Hölder continuous function $f:[0,1]^{d}\to \mathbb{R}$. $\mathcal{H}_{Q}^{\beta}\left(\mathcal{C},\mathbb{R}\right)$ is defined in a similar manner.
\end{definition}

\begin{definition}
Let $\mathcal{C}$ be the Cantor set. $\mathcal{H}_{Q}^{\beta}\left(\mathcal{C},\mathbb{R}\right)$ is defined in a manner similar to Definition \ref{holder}.
\end{definition}
\begin{proposition}[\cite{schmidt2021kolmogorov}, Theorem 2]\label{KST}
Let $d\in\mathbb{N}_{\geq1}$. There exists a monotone function $\phi:[0,1] \rightarrow \mathcal{C}$  such that for any function $f\in\mathcal{H}_{Q}^{\beta}\left([0,1]^{d},\mathbb{R}\right)$, we can find a function $g\in\mathcal{H}_{2^\beta Q}^{\frac{\beta \log 2}{d \log 3}}(\mathcal{C},\mathbb{R})$ such that
$$
f\left(x_1, \ldots, x_d\right)=g\left(3 \sum_{p=1}^d 3^{-p} \phi\left(x_p\right)\right).
$$
\end{proposition}


In \cite{schmidt2021kolmogorov}, the explicit form of $\phi(x)$ is provided. For $x=[0.a_1^xa_2^x\dots]_2\in[0,1]$,
$$
\phi(x)=\sum_{j=1}^{\infty} {2 a_j^x}3^{-1-d(j-1)}.
$$
An approximation of $\phi(x)$ with the precision parameter $K$ is given by
\begin{align*}
\phi_K(x):=\sum_{j=1}^K 2 a_j^{x} 3^{-1-d(j-1)},
\end{align*}
which will be used in our construction.







\subsection{Notations}

\begin{itemize}

    \item 

$\mathbb{N}_{\geq1}$ is the set consisting of all positive integers. $[N]=\{1,2,\dots,N\}$.

\item $\mathcal{H}_{Q}^{\beta}\left([0,1]^{d\times n},\mathbb{R}\right)$ is defined in a manner similar to Definition \ref{holder} by only 
The definition of $\mathcal{H}_{Q}^{\beta}\left([0,1]^{d\times n},\mathbb{R}\right)$ is the same as Definition \ref{holder}, except that the vector infinity norm in \eqref{holder1} is replaced by the element-wise matrix infinity norm. For $\boldsymbol{f}:[0,1]^{d\times n}\to\mathbb{R}^{d\times n}$, we say that $\boldsymbol{f}\in\mathcal{H}_{Q}^{\beta}([0,1]^{d\times n},\mathbb{R}^{d\times n})$ if $f_{rs}\in\mathcal{H}_{Q}^{\beta}([0,1]^{d\times n},\mathbb{R})$ for all $r\in[n],s\in[d]$.

\item Let $p\in[1,\infty)$. The distances $d_\infty$ and $d_p$ between two functions $\boldsymbol{f},\boldsymbol{g}\in\mathbb{R}^{d\times n}\to\mathbb{R}^{d\times n}$ are defined in the follwing manners:
\begin{align*}
d_\infty(\boldsymbol{f},\boldsymbol{g})&:=\max_{\boldsymbol{X}\in[0,1]^{d\times n}}\max_{r\in[d],s\in[n]}|{g_{rs}}(\boldsymbol{X})-f_{rs}(\boldsymbol{X})|,\\
d_p(\boldsymbol{f},\boldsymbol{g})&:=
\left(\int_{\boldsymbol{X}\in[0,1]^{d\times n}}\sum_{r\in[d],s\in[n]}|{g_{rs}}(\boldsymbol{X})-f_{rs}(\boldsymbol{X})|^pd\boldsymbol{X}\right)^{1/p}.
\end{align*}


\item Let $\sigma_R(x)=\left\{\begin{matrix}
 x, & x\geq0;\\
 0, & x<0
\end{matrix}\right.$, $\sigma_F(x)=\lfloor x \rfloor$. Let $\sigma_{RC}(x)=\frac{1}{\alpha+x}$ with $\alpha$ being any transcendental number (A transcendental number is a {complex} number that is not a root of any polynomial equation with rational coefficients). Let $\sigma_{NP}$ be any real analytic function that is nonpolynomial in some interval.

\item Let $\mathcal{FF}(L,W,A)$ be the set that contains all feedforward neural network functions with depth $L$, width $W$ and activation functions in set $A$.

\end{itemize}

\section{Transformer Can Overcome the Curse of Dimensionality}\label{main results}

\subsection{Main Results}
The following two theorems are the main results of this paper. Here $f_{\max}, g_{\max}$ and $g_{\min}$ are defined by 
\begin{align*}
f_{\max}:&=\max_{1\leq r\leq d,1\leq s\leq n}\max_{\boldsymbol{x}\in[0,1]^d}f_{rs}(\boldsymbol{x}),\\
g_{\max}:&=\max_{1\leq r\leq d,1\leq s\leq n}\max_{x\in\mathcal{C}}g_{rs}(x),\\
g_{\min}:&=\min_{1\leq r\leq d,1\leq s\leq n}\min_{x\in\mathcal{C}}g_{rs}(x),
\end{align*}
where $\{f_{rs}\}$ are the components of $\boldsymbol{f}\in\mathcal{H}_{Q}^{\beta}\left([0,1]^{d\times n},\mathbb{R}^{d\times n}\right)$, $\{g_{rs}\}$ are the outer functions in Proposition \ref{KST} for $\{f_{rs}\}$.

\begin{theorem}\label{Linfty}
Let $\boldsymbol{f}\in\mathcal{H}_{Q}^{\beta}\left([0,1]^{d\times n},\mathbb{R}^{d\times n}\right)$. For any $\epsilon>0$, there exist feedforward blocks $\boldsymbol{\mathcal{F}}_{FF}^{(sr)}:\mathbb{R}^{d\times n}\to\mathbb{R}^{2d\times n},\boldsymbol{\mathcal{\widetilde{F}}}_{FF}^{(mmr)}:\mathbb{R}^{2d\times n}\to\mathbb{R}^{d\times n}$ and a self-attention layer $\boldsymbol{\mathcal{F}}_{SA}:\mathbb{R}^{2d\times n}\to\mathbb{R}^{2d\times n}$ such that for the transformer $\boldsymbol{T}:\mathbb{R}^{d\times n}\to\mathbb{R}^{d\times n}$ defined by
\begin{align*}
\boldsymbol{T}:=\boldsymbol{\mathcal{\widetilde{F}}}_{FF}^{(mmr)}\circ\boldsymbol{\mathcal{F}}_{SA}\circ\boldsymbol{\mathcal{F}}_{FF}^{(sr)},
\end{align*}
there holds
\begin{align*}
d_\infty(\boldsymbol{T},\boldsymbol{f})
\leq\epsilon.
\end{align*}
$\boldsymbol{\mathcal{F}}_{FF}^{(sr)}$ is a feedforward block with depth $\left\lceil\frac{1}{\beta}\log_2\frac{2^{1-\beta}Q}{\epsilon}\right\rceil+3$, width $3dn$ and activation functions being ReLU and floor.  The form of $\boldsymbol{\mathcal{\widetilde{F}}}_{FF}^{(mmr)}$ can be one of the following:
\begin{enumerate}[label=(\arabic*)]
\item Its activation functions are ReLU and floor, its depth is $7dn-2$ and its width is $d\left(2\left\lceil n\left(\frac{2^{1-\beta}Q}{\epsilon}\right)^{2/\beta}\right\rceil+2\right)\left\lceil\log_2\frac{2(g_{\max}-g_{\min})}{\epsilon}\right\rceil$;
\item Its activation functions are ReLU, floor and $2^x$, its depth is $6$ and its width is $\left\lceil3d\log_2\frac{2(g_{\max}-g_{\min})}{\epsilon}\right\rceil$;
\item Its activation functions are ReLU, floor and $\sigma_{NP}$, its depth is $3$ and its width is $3d$;
\item Its activation functions are ReLU, floor and $\sigma_{RC}$, its depth is $3$ and its width is $3d$.
\end{enumerate}
\end{theorem}

\begin{theorem}\label{Lp}
Let $\boldsymbol{f}\in\mathcal{H}_{Q}^{\beta}\left([0,1]^{d\times n},\mathbb{R}^{d\times n}\right)$. Let $p\in[1,\infty)$. For any $\epsilon>0$, there exist feedforward blocks $\boldsymbol{\mathcal{F}}_{FF}^{(sr)}:\mathbb{R}^{d\times n}\to\mathbb{R}^{2d\times n},\boldsymbol{\mathcal{\widetilde{F}}}_{FF}^{(mmr)}:\mathbb{R}^{2d\times n}\to\mathbb{R}^{d\times n}$ and a self-attention layer $\boldsymbol{\mathcal{F}}_{SA}:\mathbb{R}^{2d\times n}\to\mathbb{R}^{2d\times n}$ such that for the transformer $\boldsymbol{T}:\mathbb{R}^{d\times n}\to\mathbb{R}^{d\times n}$ defined by
\begin{align*}
\boldsymbol{T}:=\boldsymbol{\mathcal{\widetilde{F}}}_{FF}^{(mmr)}\circ\boldsymbol{\mathcal{F}}_{SA}\circ\boldsymbol{\mathcal{F}}_{FF}^{(sr)},
\end{align*}
there holds
\begin{align*}
d_p(\boldsymbol{T},\boldsymbol{f})
\leq\epsilon.
\end{align*}
$\boldsymbol{\mathcal{F}}_{FF}^{(sr)}$ is a feedforward block with depth 
\begin{align*}
2\max&\left\{\left\lceil\frac{1}{\beta}\log_2\frac{2^{1/p}d^{2/p}n^{2/p}(B_{\sigma}+f_{\max})}{\epsilon}\right\rceil,
\left\lceil\frac{1}{\beta}\log_2\frac{2^{1-\beta}(2dn)^{1/p}Q}{\epsilon}\right\rceil\right\},
\end{align*}
width $4dn$ and activation functions being ReLU. The form of $\boldsymbol{\mathcal{\widetilde{F}}}_{FF}^{(mmr)}$ can be one of the following:

\begin{enumerate}[label=(\arabic*)]
\item Its activation functions are ReLU, sine and $2^x$, its depth is $4$ and its width is $2d\left\lceil\log_2\frac{2(2dn)^{1/p}(g_{\max}-g_{\min})}{\epsilon}\right\rceil$.
\item Its activation functions are ReLU, cosine and $3^x$, its depth is $5$ and its width is $d\left\lceil\log_2\frac{2(2dn)^{1/p}(g_{\max}-g_{\min})}{\epsilon}\right\rceil$.
\item Its activation functions are $\sigma_P$ and $\sigma_{NP}$, its depth is $3$ and its width is $3d$.
\item Its activation functions are $\sigma_P$ and $\sigma_{RC}$, its depth is $3$ and its width is $3d$.
\end{enumerate}
\end{theorem}

\begin{remark}
The size of $\boldsymbol{\mathcal{\widetilde{F}}}_{FF}^{(mmr)}$ presented in Theorem \ref{Linfty} gradually decreases. We can observe that the parameter number of (1) is 
$\mathcal{O}(\log\frac{1}{\epsilon})$; whereas the parameter number of (4) is $\mathcal{O}(1)$. The same applies to Theorem \ref{Lp}.
\end{remark}

\begin{remark}
Based on the proofs in Section \ref{proof}, we know that Theorems \ref{Linfty} and \ref{Lp} only present a subset of Transformers that can overcome the curse of dimensionality. Our goal here is to limit the total number of activation function types in the feedforward layers to no more than three. In fact, if more activation functions are allowed, according to Lemmas \ref{implementation of the inner matrix}, \ref{mmr-binary} and \ref{mmr-iw}, we can generate more Transformers that does not suffer from the curse of dimensionality.
\end{remark}

\subsection{Basic Idea of Our Construction}\label{basic idea}
In this section, we illustrate the basic idea of how we construct the transformer by considering
the case when $d=2, n=3$. In this case,
\begin{align*}
\boldsymbol{x}=
\begin{pmatrix}
 x_{11} & x_{12} & x_{13}\\
x_{21} & x_{22} & x_{23}
\end{pmatrix}\in\mathbb{R}^{2\times3}
\end{align*}
and the target function $\boldsymbol{f}\in\mathcal{H}_{Q}^{\beta}\left([0,1]^{2\times 3},\mathbb{R}^{2\times 3}\right)$ can be written in the following way:
\begin{align*}
\boldsymbol{f}(\boldsymbol{x})=\begin{pmatrix}
{f}_{11}(\boldsymbol{x})  & {f}_{12}(\boldsymbol{x}) & {f}_{13}(\boldsymbol{x})\\
{f}_{21}(\boldsymbol{x})  & {f}_{22}(\boldsymbol{x}) & {f}_{23}(\boldsymbol{x})
\end{pmatrix},
\end{align*}
where $f_{rs}\in\mathcal{H}_{Q}^{\beta}\left([0,1]^{2\times 3},\mathbb{R}\right)$ for $r=1,2;s=1,2,3$.

Proposition \ref{KST}, the Kolmogorov–Arnold representation theorem, enables us to find $g_{rs}\in\mathcal{H}_{2^\beta Q}^{\frac{\beta \log 2}{6 \log 3}}(\mathcal{C},\mathbb{R})$ such that for $r=1,2;s=1,2,3$,
\begin{align*}
f_{rs}(\boldsymbol{x})=g_{rs}\left(3\sum_{p=1}^{2}\sum_{q=1}^3a_{pq}\phi(x_{pq})\right).
\end{align*}
where $a_{pq}:=\frac{1}{3^{3(p-1)+q}}$.

Let $\phi_K(x)$ be an approximant of $\phi(x)$:
\begin{align*}
\phi_K(x):=2\sum_{j=1}^Ka_{j}^x3^{-6(j-1)-1}
\end{align*}
and
\begin{align*}
a(\boldsymbol{x})&:=\sum_{p=1}^2\sum_{q=1}^3a_{pq}\phi_K(x_{pq}),\\
b(\boldsymbol{x})&:=3^{6K+1}a(\boldsymbol{x})+1.
\end{align*}
It can be seen that $g_{rs}\left(3a(\boldsymbol{x})\right)$ is an approximant of $f_{rs}(\boldsymbol{x})$.

Below is a rough process of our construction in this paper. Here, steps one to two are implemented using a feedforward block; step three is implemented using a self-attention layer; steps four to six are implemented using another feedforward block.
\begin{align*}
&\begin{pmatrix}
 x_{11} & x_{12} & x_{13}\\
x_{21} & x_{22} & x_{23}
\end{pmatrix}\\
&\Longrightarrow 
\begin{pmatrix}
 x_{11} & x_{12}+2 & x_{13}+4\\
x_{21} & x_{22}+2 & x_{23}+4
\end{pmatrix}\\
&\Longrightarrow
\begin{pmatrix}
 \sum_{p}a_{p1}\phi_K(x_{p1}) & \sum_{p}a_{p2}\phi_K(x_{p2}) & \sum_{p}a_{p3}\phi_K(x_{p3})\\
 \sum_{p}a_{p1}\phi_K(x_{p1}) & \sum_{p}a_{p2}\phi_K(x_{p2}) & \sum_{p}a_{p3}\phi_K(x_{p3})
\end{pmatrix}\\
&\Longrightarrow\begin{pmatrix}
 a(\boldsymbol{x}) & a(\boldsymbol{x}) & a(\boldsymbol{x})\\
 a(\boldsymbol{x}) & a(\boldsymbol{x}) & a(\boldsymbol{x})
\end{pmatrix}\\
&\Longrightarrow\begin{pmatrix}
 b(\boldsymbol{x}) & b(\boldsymbol{x}) & b(\boldsymbol{x})\\
 b(\boldsymbol{x}) & b(\boldsymbol{x}) & b(\boldsymbol{x})
\end{pmatrix}\\
&\Longrightarrow\begin{pmatrix}
 b(\boldsymbol{x}) & b(\boldsymbol{x})+3^{6K} & b(\boldsymbol{x})+2\cdot3^{6K}\\
 b(\boldsymbol{x}) & b(\boldsymbol{x})+3^{6K} & b(\boldsymbol{x})+2\cdot3^{6K}
\end{pmatrix}\\
&\Longrightarrow\begin{pmatrix}
 g_{11}(3a(\boldsymbol{x})) & g_{12}(3a(\boldsymbol{x})) & g_{13}(3a(\boldsymbol{x}))\\
 g_{21}(3a(\boldsymbol{x})) & g_{22}(3a(\boldsymbol{x})) & g_{23}(3a(\boldsymbol{x}))
\end{pmatrix}
\end{align*}
Now let's explain each of these steps. First, we are going to explain steps one and two. It is difficult to directly transform $\boldsymbol{x}$ into the matrix in the {third} row. This is because to achieve this, it is necessary to use three different column functions to map each column (a two-dimensional vector) to a new two-dimensional vector. Note that these column functions are each unique pairwise, as can be understood from the fact that the combination coefficients for the first column are $a_{p1}$ while for the second column they are $a_{p2}$. Since $\phi_K$ can be (approximately) implemented using a feedforward neural network function, each column function can also be (approximately) implemented using a feedforward neural network function. However, because the three column functions are different, it is not possible to use the same feedforward neural network function to simultaneously implement the transformation of all three columns. To address this issue, we have developed a technique here: by applying different translations to the three columns, at the cost of broadening the domain of the column functions, we transform the problem of implementing three distinct column functions into that of implementing a single column function. Therefore, the role of step one is to perform a translation on each column, while step two involves implementing a specific column function.

Step three involves summing the matrix columns, an operation that can be implemented using a self-attention layer.

Step four is a linear mapping step designed to ensure that the points to be memorized in the memory task in step six are all integers. This allows us to utilize the results of integer memorization from the feedforward neural network (Lemma \ref{binary-app},\ref{NP-ri},\ref{RC-ri}). Step five is another translation step, with a purpose similar to the first step: through this translation step, we transform the problem of implementing three different column functions into implementing a single column function. Step six involves two memorization problems concerning up to $3 \cdot 3^{6K}$ integers:
given $r\in\{1,2\}$, find a feedforward neural network function $f_{FF,r}^{(mmr)}$ such that for any $\boldsymbol{x}\in[0,1]^d$ and $s\in\{1,2,3\}$, there holds
\begin{align*}
f_{FF,r}^{(mmr)}\left(b(\boldsymbol{x})+(s-1)\cdot3^{6K}\right)=g_{rs}\left(3a(\boldsymbol{x})\right).
\end{align*}
This is because the set
\begin{align*}
\Lambda:=\{b(\boldsymbol{x})+(s-1)\cdot3^{6K}:\boldsymbol{x}\in[0,1]^{2\times3},s\in\{1,2,3\}\}
\end{align*}
is finite and 
$\Lambda\subset\left\{1,2,\dots,3\cdot3^{6K}\right\}$.

\begin{remark}
The cardinality of $\Lambda$ is actually 
$|\Lambda|=3\cdot2^{6K}$, which we here magnify to $3\cdot3^{6K}$. This manipulation has a negligible effect on the orders of the size of the transformers in Theorems \ref{Linfty} and \ref{Lp}.
\end{remark}

\begin{remark}
From the process described above, we can see that the summation required in the KST is divided into two parts in its construction. The summation over rows is achieved by the feedforward layers, while the summation over columns is accomplished by the self-attention mechanism.
\end{remark}

\subsection{Proof of Theorems \ref{Linfty} and \ref{Lp}}\label{proof}

We define the inner matrix as the matrix in step five in section \ref{basic idea}. More generally, given
\begin{align*}
\boldsymbol{X}=
\begin{pmatrix}
x_{11}  & x_{12} & \cdots & x_{1n}\\
x_{21}  & x_{22} & \cdots & x_{2n}\\
\vdots  & \vdots & \ddots & \vdots\\
x_{d1}  & x_{d2} & \cdots & x_{dn}
\end{pmatrix}\in[0,1]^{d\times n},
\end{align*}
define the inner matrix $\boldsymbol{Z}:=\boldsymbol{Z}(\boldsymbol{X})\in\mathbb{R}^{d\times n}$ by
\begin{align*}
&\boldsymbol{Z}_{:,s}=
\left[\sum_{q=1}^n\sum_{p=1}^{d}\frac{3^{Kdn+1}}{3^{(p-1)s+q}}\phi_K(x_{pq})+1+(s-1)\cdot3^{Kdn}\right]\boldsymbol{1}_{d\times 1} 
\end{align*}
for $s\in[n]$. With this definition, the proof of Theorems \ref{Linfty} and \ref{Lp} can be divided into two parts: one part proves that for (almost) all $\boldsymbol{X}\in[0,1]^{d\times n}$, there exist a feedforward block $\boldsymbol{\mathcal{F}}_{FF}^{(sr)}$, a self-attention layer $\boldsymbol{\mathcal{F}}_{SA}$, and a linear transformation layer $\boldsymbol{\mathcal{A}}$ such that 
\begin{align*}
\boldsymbol{Z}=\boldsymbol{\mathcal{A}}\circ\boldsymbol{\mathcal{F}}_{SA}\circ\boldsymbol{\mathcal{F}}_{FF}^{(sr)}(\boldsymbol{X});
\end{align*}
the other part proves that for a given $\boldsymbol{Z}$, there exists a feedforward block 
$\boldsymbol{\mathcal{F}}_{FF}^{(mmr)}$ such that
\begin{align*}
\boldsymbol{\mathcal{F}}_{FF}^{(mmr)}(\boldsymbol{Z})\approx \boldsymbol{f}(\boldsymbol{X}).
\end{align*}
These two parts correspond respectively to steps one to five and step six in section \ref{basic idea}. The first part is guaranteed by Lemma \ref{implementation of the inner matrix} below.

\begin{lemma}\label{implementation of the inner matrix}
Let $K\in\mathbb{N}_{\geq1}$. Let $p\in[1,\infty)$. There exist a feedforward block $\boldsymbol{\mathcal{F}}_{FF}^{(sr)}:\mathbb{R}^{d\times n}\to\mathbb{R}^{2d\times n}$, a self-attention layer $\boldsymbol{\mathcal{F}}_{SA}:\mathbb{R}^{2d\times n}\to\mathbb{R}^{2d\times n}$ and a linear transformation layer $\boldsymbol{\mathcal{A}}:\mathbb{R}^{2d\times n}\to\mathbb{R}^{d\times n}$ such that
\begin{align}\label{implementation of the inner matrix1}
\boldsymbol{Z}=\boldsymbol{\mathcal{A}}\circ\boldsymbol{\mathcal{F}}_{SA}\circ\boldsymbol{\mathcal{F}}_{FF}^{(sr)}(\boldsymbol{X}).
\end{align}
The form of $\boldsymbol{\mathcal{F}}_{FF}^{(sr)}$ can be one of the following:
\begin{enumerate}[label=(\arabic*)]
\item Its activation function is ReLU, its depth is $2K$ and its width is $4dn$. 
\item Its activation functions are ReLU and floor, its depth is $K+3$ and its width is $3dn$. 
\end{enumerate}
For the first case, (\ref{implementation of the inner matrix1}) holds for $
\boldsymbol{X}\in\{[0,1]\backslash\Omega^{(flow)}\}^{d\times n}
$; for the second case, (\ref{implementation of the inner matrix1}) holds for $\boldsymbol{X}\in[0,1]^{d\times n}$. Here $\Omega^{(flaw)}\subset[0,1]$ is some region with the Lebesgue measure not greater than $2^{-K\beta p}$.
\end{lemma}

The second part is guaranteed by Lemmas \ref{mmr-binary} and \ref{mmr-iw} below, where the former is based on binary memorization, and the latter is based on the memorization with labels of arbitrary real numbers.

\begin{lemma}\label{mmr-binary}
Let $K,H\in\mathbb{N}_{\geq1}$. There exists a feedforward neural network function $\boldsymbol{f}_{FF}^{(mmr)}:\mathbb{R}^d\to\mathbb{R}^d$ such that for $\boldsymbol{\widetilde{f}}(\boldsymbol{X}):=\boldsymbol{\mathcal{F}}_{FF}^{(mmr)}(\boldsymbol{Z}(\boldsymbol{X}))$, there holds
\begin{align*}
|\widetilde{f}_{rs}(\boldsymbol{X})-{f}_{rs}(\boldsymbol{X})|\leq\frac{g_{\max}-g_{\min}}{2^H}+\frac{2^{\beta}Q}{2^{(K+2)\beta}},
\end{align*}
for any $\boldsymbol{X}\in[0,1]^d,r\in[d],s\in[n]$. $\boldsymbol{f}_{FF}^{(mmr)}$ can be in one of the following feedforward neural network classes:
\begin{enumerate}[label=(\arabic*)]
\item $\mathcal{FF}\left(7L-2,\left(2\left\lceil (n3^{Kdn})^{1/L}\right\rceil+2\right)dH,\{\text{ReLU, floor}\}\right)$,
where $L$ can be any positive integer number;
\item $\mathcal{FF}(6,3dH,\{\text{ReLU, floor}, 2^x\})$;
\item $\mathcal{FF}(4,2dH,\{\text{ReLU, sine}, 2^x\})$;
\item $\mathcal{FF}(5,dH,\{\text{ReLU, cosine}, 3^x\})$.
\end{enumerate}
\end{lemma}

\begin{lemma}\label{mmr-iw}
Let $K\in\mathbb{N}_{\geq1}$. There exists a feedforward neural network function $\boldsymbol{f}_{FF}^{(mmr)}:\mathbb{R}^d\to\mathbb{R}^d$ such that for $\boldsymbol{\widetilde{f}}(\boldsymbol{X}):=\boldsymbol{\mathcal{F}}_{FF}^{(mmr)}(\boldsymbol{Z}(\boldsymbol{X}))$, there holds
\begin{align*}
|\widetilde{f}_{rs}(\boldsymbol{X})-{f}_{rs}(\boldsymbol{X})|\leq\delta+\frac{2^{\beta}Q}{2^{(K+2)\beta}},
\end{align*}
for any $\boldsymbol{X}\in[0,1]^d,r\in[d],s\in[n]$. Here $\delta>0$ can be arbitrarily small. $\boldsymbol{f}_{FF}^{(mmr)}$ can be in one of the following feedforward neural network classes:
\begin{enumerate}[label=(\arabic*)]
\item $\mathcal{FF}(3,d,\{\sigma_P,\sigma_{NP}\})$;
\item $\mathcal{FF}(3,d,\{\sigma_P,\sigma_{RC}\})$.
\end{enumerate}
\end{lemma}

\subsubsection{Proof of Theorem \ref{Linfty}}
(1) and (2) arise from (2) of Lemma \ref{implementation of the inner matrix} and (1)(2) of Lemma \ref{mmr-binary}. (3) and (4) arise from (2) of Lemma \ref{implementation of the inner matrix} and (1)(2) of Lemma \ref{mmr-iw}. Note that here the general periodic function $\sigma_P(x)$ in Lemma \ref{mmr-iw} is replaced by a specific periodic function with period $1$ and equaling to $x$ on $[0,1)$. This function can be implemented by a ReLU-floor feedforward network function with depth $2$ and width $3$:
\begin{align*}
\sigma_R(x)+\sigma_R(-x)-\sigma_F(x).
\end{align*}
$\boldsymbol{\mathcal{\widetilde{F}}}_{FF}^{(mmr)}$ is defined by
\begin{align*}
\boldsymbol{\mathcal{\widetilde{F}}}_{FF}^{(mmr)}:=\boldsymbol{\mathcal{{F}}}_{FF}^{(mmr)}\circ\boldsymbol{\mathcal{A}},
\end{align*}
where $\boldsymbol{\mathcal{A}}$ is from Lemma \ref{implementation of the inner matrix} and $\boldsymbol{\mathcal{{F}}}_{FF}^{(mmr)}$ is from Lemma \ref{mmr-binary} or Lemma \ref{mmr-iw}. Parameters in these lemmas are set to be
\begin{align*}
H&=\left\lceil\log_2\frac{2(g_{\max}-g_{\min})}{\epsilon}\right\rceil,\\
K&=\left\lceil\frac{1}{\beta}\log_2\frac{2^{1-\beta}Q}{\epsilon}\right\rceil,\\
L&=dn.
\end{align*}


\subsubsection{Proof of Theorem \ref{Lp}}
(1) and (2) arise from (1) of Lemma \ref{implementation of the inner matrix} and (3)(4) of Lemma \ref{mmr-binary}. (3) and (4) arise from (1) of Lemma \ref{implementation of the inner matrix} and (1)(2) of Lemma \ref{mmr-iw}. $\boldsymbol{\mathcal{\widetilde{F}}}_{FF}^{(mmr)}$ is defined by
\begin{align*}
\boldsymbol{\mathcal{\widetilde{F}}}_{FF}^{(mmr)}:=\boldsymbol{\mathcal{{F}}}_{FF}^{(mmr)}\circ\boldsymbol{\mathcal{A}},
\end{align*}
where $\boldsymbol{\mathcal{A}}$ is from Lemma \ref{implementation of the inner matrix} and $\boldsymbol{\mathcal{{F}}}_{FF}^{(mmr)}$ is from Lemma \ref{mmr-binary} or Lemma \ref{mmr-iw}.

To bound the $L^p$ distance of $\boldsymbol{T}$ and $\boldsymbol{f}$, we decompose the whole integral into two parts: 
\begin{align*}
&[d_p(\boldsymbol{T},\boldsymbol{f})]^p\\
&=\int_{\boldsymbol{X}\in[0,1]^d}\sum_{r\in[d],s\in[n]}|T_{rs}(\boldsymbol{X})-f_{rs}(\boldsymbol{X})|^pd\boldsymbol{X}\\
&=\int_{\boldsymbol{X}\in\Omega_1
}\sum_{r\in[d],s\in[n]}|T_{rs}(\boldsymbol{X})-f_{rs}(\boldsymbol{X})|^pd\boldsymbol{X}
+\int_{\boldsymbol{X}\in\Omega_2
}\sum_{r\in[d],s\in[n]}|T_{rs}(\boldsymbol{X})-f_{rs}(\boldsymbol{X})|^pd\boldsymbol{X},
\end{align*}
where $\Omega_1 := \{[0,1] \setminus \Omega^{(flow)}\}^{d \times n}$ denotes the region where $T_{rs}$ can effectively approximate $f_{rs}$, whereas $\Omega_2 := [0,1]^{d \times n} \setminus \{[0,1] \setminus \Omega^{(flow)}\}^{d \times n}$ represents the region where $T_{rs}$ fails to approximate $f_{rs}$ well.

Lemma \ref{mmr-binary} enables us to derive an upper bound for the first term(the cases when adopting Lemma \ref{mmr-iw} can be handled almost the same and hence we omit this part):
\begin{align*}
dn\left(\frac{g_{\max}-g_{\min}}{2^H}+\frac{2^{\beta}Q}{2^{(K+2)\beta}}\right)^p.
\end{align*}

By our construction (see Lemma \ref{mmr-binary} and Lemma \ref{mmr-iw} for details), the range of $T_{rs}$ is bounded by some constants $B_{\sigma}$, therefore the integrand of the second term is simply bounded by $dn(B_{\sigma}+f_{\max})^p$. Furthermore, since $m\left(\Omega^{(flow)}\right)\leq2^{-K\beta p}$, by Bernoulli's inequality we can derive a bound for the measure of the region of the second term: 
\begin{align*}
m\left([0,1]^d\backslash\{[0,1]\backslash\Omega^{(flow)}\}^{d\times n}\right)
&=1-\left(1-m\left(\Omega^{(flow)}\right)\right)^{dn}\\
&\leq dn\cdot m\left(\Omega^{(flow)}\right)\leq dn2^{-K\beta p}.
\end{align*}
Therefore the second term is bounded by 
\begin{align*}
2^{-K\beta p}d^2n^2(B_{\sigma}+f_{\max})^p.
\end{align*}
By far we obtain
\begin{align*}
&[d_p(\boldsymbol{T},\boldsymbol{f})]^p
\leq
dn\left(\frac{g_{\max}-g_{\min}}{2^H}+\frac{2^{\beta}Q}{2^{(K+2)\beta}}\right)^p
+\frac{d^2n^2(B_{\sigma}+f_{\max})^p}{2^{K\beta p}}.
\end{align*}
By setting the parameters
\begin{align*}
K&=\max\left\{\left\lceil\frac{1}{\beta}\log_2\frac{2^{1/p}d^{2/p}n^{2/p}(B_{\sigma}+f_{\max})}{\epsilon}\right\rceil,\left\lceil\frac{1}{\beta}\log_2\frac{2^{1-\beta}(2dn)^{1/p}Q}{\epsilon}\right\rceil\right\},\\
H&=\left\lceil\log_2\frac{2(2dn)^{1/p}(g_{\max}-g_{\min})}{\epsilon}\right\rceil,
\end{align*}
we arrive the desired accuracy $\epsilon$.

\section{Conclusions}\label{Conclusions}

This paper investigates the approximation of the Hölder continuous function class by Transformers and constructs several Transformers that can overcome the curse of dimensionality. These Transformers consist of one self-attention layer with one head and the softmax function as the activation function, along with several small-size feedforward layers. These results demonstrate that Transformers have a strong expressive capability. The construction in this paper is based on the Kolmogorov-Arnold Representation Theorem and does not require the concept of contextual mapping. Additionally, the translation technique proposed in this paper helps to apply the previous approximation results of feedforward neural networks to transformer research.

There are several potential future research directions. Further exploration of the expressive power of Transformers using KST is a direction worth considering. Investigating how Transformers approximate functions in other function spaces, such as \( C^s \) spaces, Sobolev spaces, etc., poses an intriguing question. Integrating the approximation issue with the specific tasks faced by Transformers is also a meaningful subject to explore.

\bibliographystyle{plainnat}
\bibliography{ref}

\appendix

\section{Proof of Lemma \ref{implementation of the inner matrix}}
We only present a proof for the ReLU case. The proof for the floor-ReLU case only differs subtly and we will point out these differences in the proof. 

Let $f_{FF}^{(R-pi)}$ be the ReLU-feedforward network function in Lemma \ref{Rpi}(For the floor-ReLU case, we make use of Lemma \ref{FRpi}). Define the
multi-dimensional piecewise inner function $\boldsymbol{f}_{FF}^{(mpi)}:\mathbb{R}^d\to\mathbb{R}^{2d}$ by
\begin{align*}
\boldsymbol{f}_{FF}^{(mpi)}(\boldsymbol{x}):&=
\begin{pmatrix}
\frac{1}{3}\boldsymbol{1}_{d\times1}  & \frac{1}{3^{n+1}}\boldsymbol{1}_{d\times1} & \cdots & \frac{1}{3^{(d-1)n+1}}\boldsymbol{1}_{d\times1}\\
\boldsymbol{0}_{d\times1}  & \boldsymbol{0}_{d\times1} & \cdots & \boldsymbol{0}_{d\times1}
\end{pmatrix}
\begin{pmatrix}
f_{FF}^{(R-pi)}(x_1) \\
f_{FF}^{(R-pi)}(x_2) \\
\vdots\\
f_{FF}^{(R-pi)}(x_d)
\end{pmatrix}\\
&=\begin{pmatrix}
\left[\sum_{p=1}^{d}\frac{1}{3^{(p-1)n+1}}f_{FF}^{(R-pi)}(x_p)\right]\boldsymbol{1}_{d\times 1} \\
\boldsymbol{0}_{d\times 1}
\end{pmatrix}.
\end{align*}
By Lemma \ref{Rpi}, it follows that for $\boldsymbol{x}\in\{[2q,2q+1]\backslash(\Omega^{(flow)}+2q)\}^d,q\in\{0,1,\dots,n-1\}$, there holds
\begin{align*}
\boldsymbol{f}_{FF}^{(mpi)}(\boldsymbol{x})=
\begin{pmatrix}
\left[\frac{1}{3^{q-1}}\sum_{p=1}^{d}\frac{1}{3^{(p-1)n+1}}\phi_K(x_p-2q)+\frac{(3^q-1)(3^{dn}-1)}{2\cdot3^{q-1}(3^{dn}-3^{(d-1)n})}\right]\boldsymbol{1}_{d\times 1} \\
\boldsymbol{0}_{d\times 1}
\end{pmatrix}.
\end{align*}
Let
\begin{align*}
\boldsymbol{b}_0:=\begin{pmatrix}
0\cdot\boldsymbol{1}_{d\times 1}  & 2\cdot\boldsymbol{1}_{d\times 1} & \cdots & 2(n-1)\cdot\boldsymbol{1}_{d\times 1}
\end{pmatrix}
\in\mathbb{R}^{d\times n}. 
\end{align*}
Define $\boldsymbol{Z}_1:=\boldsymbol{\mathcal{F}}_{FF}^{(mpi)}(\boldsymbol{X}+\boldsymbol{b}_0)\in\mathbb{R}^{2d\times n}$. For
$
\boldsymbol{X}_{:,s}\in\{[0,1]\backslash\Omega^{(flow)}\}^d
$,
$
\boldsymbol{X}_{:,s}+(\boldsymbol{b}_0)_{:,s}\in\{[2q,2q+1]\backslash\Omega^{(flow)}\}^d
$. Therefore it follows that
\begin{align*}
\boldsymbol{Z}_1=\begin{pmatrix}
\left[3\sum_{p=1}^{d}\frac{1}{3^{(p-1)n+1}}\phi_K(x_{p1}-2q)+\frac{(3^0-1)(3^{dn}-1)}{2\cdot3^{-1}(3^{dn}-3^{(d-1)n})}\right]\boldsymbol{1}_{1\times d}  & \boldsymbol{0}_{1\times d}\\
\left[3\sum_{p=1}^{d}\frac{1}{3^{(p-1)n+2}}\phi_K(x_{p2}-2q)+\frac{(3^1-1)(3^{dn}-1)}{2\cdot3^{0-1}(3^{dn}-3^{(d-1)n})}\right]\boldsymbol{1}_{1\times d}  & \boldsymbol{0}_{1\times d}\\
\vdots  & \vdots\\
\left[3\sum_{p=1}^{d}\frac{1}{3^{(p-1)n+n}}\phi_K(x_{pn}-2q)+\frac{(3^{n-1}-1)(3^{dn}-1)}{2\cdot3^{n-2}(3^{dn}-3^{(d-1)n})}\right]\boldsymbol{1}_{1\times d}  & \boldsymbol{0}_{1\times d}
\end{pmatrix}^T.
\end{align*}
Let 
\begin{align*}
\boldsymbol{b}_1:=\begin{pmatrix}
\frac{-(3^0-1)(3^{dn}-1)}{2\cdot3^{-1}(3^{dn}-3^{(d-1)n})}\boldsymbol{1}_{d\times 1}  & \frac{-(3^1-1)(3^{dn}-1)}{2\cdot3^{0-1}(3^{dn}-3^{(d-1)n})}\boldsymbol{1}_{d\times 1} & \cdots & \frac{-(3^{n-1}-1)(3^{dn}-1)}{2\cdot3^{n-2}(3^{dn}-3^{(d-1)n})}\boldsymbol{1}_{d\times 1}\\
\boldsymbol{0}_{d\times 1}  & \boldsymbol{0}_{d\times 1} & \cdots & \boldsymbol{0}_{d\times 1}
\end{pmatrix}\in\mathbb{R}^{2d\times n}.
\end{align*}
For the floor-ReLU case, we simply set $\boldsymbol{b}_1=\boldsymbol{0}_{2d\times n}$.
We then have
\begin{align*}
&\boldsymbol{Z}_2:=\boldsymbol{Z}_1+\boldsymbol{b}_1=\\
&\resizebox{\textwidth}{!}{$\begin{pmatrix}
3\sum_{p=1}^{d}\frac{1}{3^{(p-1)n+1}}\phi_K(x_{p1})\boldsymbol{1}_{d\times1}  & 3\sum_{p=1}^{d}\frac{1}{3^{(p-1)n+2}}\phi_K(x_{p2})\boldsymbol{1}_{d\times1} & \cdots & 3\sum_{p=1}^{d}\frac{1}{3^{(p-1)n+n}}\phi_K(x_{pn})\boldsymbol{1}_{d\times1}\\\boldsymbol{0}_{d\times 1}  & \boldsymbol{0}_{d\times 1} & \cdots & \boldsymbol{0}_{d\times 1}
\end{pmatrix}\in\mathbb{R}^{2d\times n}.$}
\end{align*}
Next we define the matrices in the self attention layer:
\begin{align*}
\boldsymbol{W}_{O}:=\begin{pmatrix}
\boldsymbol{0}_{d\times d}&\boldsymbol{0}_{d\times d}\\
\boldsymbol{I}_{d\times d}&\boldsymbol{0}_{d\times d}
\end{pmatrix},\quad\boldsymbol{W}_{V}:=\boldsymbol{I}_{2d\times 2d},\quad\boldsymbol{W}_{K}:=\boldsymbol{0}_{2d\times 2d},\quad\boldsymbol{W}_{Q}:=\boldsymbol{0}_{2d\times 2d}.
\end{align*}
Then
\begin{align*}
&\boldsymbol{Z}_3:=\boldsymbol{\mathcal{F}}_{SA}(\boldsymbol{Z}_2)
=\boldsymbol{Z}_2+\boldsymbol{W}_{O}\boldsymbol{W}_{V}\boldsymbol{Z}_2\sigma_S((\boldsymbol{W}_{V}\boldsymbol{Z}_2)^T(\boldsymbol{W}_{K}\boldsymbol{Z}_2))=\\
&\resizebox{\textwidth}{!}{$
\begin{pmatrix}
3\sum_{p=1}^{d}\frac{1}{3^{(p-1)n+1}}\phi_K(x_{p1})\boldsymbol{1}_{d\times1}  & 3\sum_{p=1}^{d}\frac{1}{3^{(p-1)n+2}}\phi_K(x_{p2})\boldsymbol{1}_{d\times1} & \cdots & 3\sum_{p=1}^{d}\frac{1}{3^{(p-1)n+n}}\phi_K(x_{pn})\boldsymbol{1}_{d\times1}\\
3\sum_{q=1}^n\sum_{p=1}^{d}\frac{1}{3^{(p-1)n+q}}\phi_K(x_{pq})\boldsymbol{1}_{d\times1} & 3\sum_{q=1}^n\sum_{p=1}^{d}\frac{1}{3^{(p-1)n+q}}\phi_K(x_{pq})\boldsymbol{1}_{d\times1} & \cdots & 3\sum_{q=1}^n\sum_{p=1}^{d}\frac{1}{3^{(p-1)n+q}}\phi_K(x_{pq})\boldsymbol{1}_{d\times1}
\end{pmatrix}.$}
\end{align*}
Define the scaling matrix
\begin{align*}
\boldsymbol{W}^{(scl)}:=\begin{pmatrix}
\boldsymbol{0}_{d\times d}  & 3^{Kdn}\boldsymbol{I}_{d\times d}
\end{pmatrix}
\in\mathbb{R}^{d\times 2d}
\end{align*}
and the segmentation matrix
\begin{align*}
\boldsymbol{b}^{(sgt)}:=\begin{pmatrix}
\boldsymbol{1}_{d\times1}  & (1+3^{Kdn})\boldsymbol{1}_{d\times1} & \cdots & (1+(n-1)\cdot3^{Kdn})\boldsymbol{1}_{d\times1}
\end{pmatrix}
\in\mathbb{R}^{d\times n}.
\end{align*}
Then $\boldsymbol{W}^{(scl)}\boldsymbol{Z}_{3}+\boldsymbol{b}^{(sgt)}$ turns out to be the inner matrix $\boldsymbol{Z}$.

\section{Proof of Lemma \ref{mmr-binary}}

Before we perform a binary expansion of $g_{rs}$, we need to normalize it so that its range lies in $[0,1]$. The normalization procedure is as follows:
\begin{align*}
g_{rs}^{(nml)}(x):=\frac{g_{rs}(x)-g_{\min}}{g_{\max}-g_{\min}},\quad r\in[d],s\in[n].
\end{align*}
Let
\begin{align*}
\Lambda:=\left\{3^{Kdn+1}\sum_{q=1}^n\sum_{p=1}^{d}\frac{1}{3^{(p-1)n+q}}\phi_K(x_{pq})+1+(s-1)\cdot3^{Kdn}:\boldsymbol{x}\in[0,1]^{d\times n},s\in[n]\right\}.
\end{align*}
Now, given $r\in[n]$, for each 
$$m=3^{Kdn+1}\sum_{q=1}^n\sum_{p=1}^{d}\frac{1}{3^{(p-1)n+q}}\phi_K(x_{pq})+1+(s-1)\cdot3^{Kdn}\in\Lambda,$$ 
we perform a binary expansion of $g_{rs}^{(nml)}$ at the point $\frac{m-1-(s-1)\cdot3^{Kdn}}{3^{Kdn}}$:
\begin{align*}
g_{rs}^{(nml)}\left(\frac{m-1-(s-1)\cdot3^{Kdn}}{3^{Kdn}}\right)=\sum_{i=1}^{\infty}\theta_{rsi}^{(m)}2^{-i},
\end{align*}
where $\theta_{rsi}^{(m)}\in\{0,1\}$. For each $r\in[d]$ and $i\in[H]$, by Lemma \ref{binary-app}, there exists a meomerization neural network function  $f_{FF,ri}^{(mmr)}(x)$ such that
\begin{align*}
f_{FF,ri}^{(mmr)}(m)=\left\{\begin{matrix}
\theta_{rsi}^{(m)},  & m\in\Lambda;\\
0,  & \text{else}.
\end{matrix}\right.
\end{align*}
Here the number of points to be memorized is $M=n\cdot3^{Kdn}$. 

Arrange $\{f_{FF,ri}^{(mmr)}\}_{r\in[d],i\in[H]}$ in parallel to form a function from $\mathbb{R}^d$ to $\mathbb{R}^{dH}$:
\begin{align*}
&\boldsymbol{f}_{FF}^{(mmr-prl)}(\boldsymbol{x}):=\\
&\resizebox{\textwidth}{!}{$\begin{pmatrix}
f_{FF,11}^{(mmr)}(x_1) & \cdots & f_{FF,1H}^{(mmr)}(x_1) & f_{NN,21}^{(mmr)}(x_2) & \cdots & f_{FF,2H}^{(mmr)}(x_2) & \cdots & f_{FF,d1}^{(mmr)}(x_d) & \cdots & f_{FF,dH}^{(mmr)}(x_d)
\end{pmatrix}^T.$}
\end{align*}
The feedforward neural network function $\boldsymbol{f}_{FF}^{(pre-mmr)}$ that implement the approximate binary expansion is then defined by
\begin{align*}
\boldsymbol{f}_{FF}^{(pre-mmr)}(\boldsymbol{x}):=\boldsymbol{W}^{(coe)}\boldsymbol{f}_{FF}^{(mmr-prl)}(\boldsymbol{x})
=\sum_{i=1}^{H}2^{-i}
\begin{pmatrix}
f_{FF,1i}^{(mmr)}(x_1) \\
f_{FF,2i}^{(mmr)}(x_2) \\
\vdots \\
f_{FF,di}^{(mmr)}(x_d)
\end{pmatrix},
\end{align*}
where the binary expansion coefficient matrix
\begin{align*}
\boldsymbol{W}^{(coe)}:=\begin{pmatrix}
\boldsymbol{w} &  &  & \\
  & \boldsymbol{w} &  & \\
  &  & \ddots & \\
  &  &  & \boldsymbol{w}
\end{pmatrix}
\in\mathbb{R}^{d\times dH}
\end{align*}
with
\begin{align*}
\boldsymbol{w}=\begin{pmatrix}
\frac{1}{2}  & \frac{1}{4} & \cdots & \frac{1}{2^H}
\end{pmatrix}
\in\mathbb{R}^{1\times H}.
\end{align*}
The feedforward function 
$\boldsymbol{{f}}_{FF}^{(mmr)}(\boldsymbol{x})$ that we are looking for is now defined by
\begin{align*}
\boldsymbol{{f}}_{FF}^{(mmr)}(\boldsymbol{x}):=\boldsymbol{W}^{(nml)}\boldsymbol{f}_{FF}^{(pre-mmr)}(\boldsymbol{x})+\boldsymbol{b}^{(nml)}
\end{align*}
where
\begin{align*}
\boldsymbol{W}^{(nml)}:=(g_{\max}-g_{\min})\boldsymbol{I}_{d\times d}\in\mathbb{R}^{d\times d},
\quad\boldsymbol{b}^{(nml)}:=g_{\min}\boldsymbol{1}_{d\times n}\in\mathbb{R}^{d\times n}.
\end{align*}
If follows that
\begin{align*}
\boldsymbol{\widetilde{f}}_{:,s}&=\boldsymbol{W}^{(nml)}\boldsymbol{f}_{FF}^{(pre-mmr)}(\boldsymbol{Z}_{:,s})+\boldsymbol{b}^{(nml)}_{:,s}\\
&=
\begin{pmatrix}
(g_{\max}-g_{\min})\sum_{i=1}^{H}2^{-i}f_{FF,1i}^{(mmr)}(Z_{1s})+g_{\min} \\
(g_{\max}-g_{\min})\sum_{i=1}^{H}2^{-i}f_{FF,2i}^{(mmr)}(Z_{2s})+g_{\min} \\
\vdots \\
(g_{\max}-g_{\min})\sum_{i=1}^{H}2^{-i}f_{FF,di}^{(mmr)}(Z_{ds})+g_{\min}
\end{pmatrix}\\
&=\begin{pmatrix}
(g_{\max}-g_{\min})\sum_{i=1}^{H}2^{-i}\theta_{1si}^{(Z_{1s})}+g_{\min} \\
(g_{\max}-g_{\min})\sum_{i=1}^{H}2^{-i}\theta_{2si}^{(Z_{2s})}+g_{\min} \\
\vdots \\
(g_{\max}-g_{\min})\sum_{i=1}^{H}2^{-i}\theta_{dsi}^{(Z_{ds})}+g_{\min}
\end{pmatrix}.
\end{align*}
Recall that
\begin{align*}
f_{rs}\left(\boldsymbol{X}\right)=g_{rs}\left(3\sum_{q=1}^n\sum_{p=1}^{d}\frac{1}{3^{(p-1)n+q}}\phi(x_{pq})\right)
\end{align*}
and make use of the following equality:
\begin{align*}
&g_{rs}\left(3\sum_{q=1}^n\sum_{p=1}^{d}\frac{1}{3^{(p-1)n+q}}\phi_K(x_{pq})\right)\\
&=(g_{\max}-g_{\min})g_{rs}^{(nml)}\left(3\sum_{q=1}^n\sum_{p=1}^{d}\frac{1}{3^{(p-1)n+q}}\phi_K(x_{pq})\right)+g_{\min}\\
&=(g_{\max}-g_{\min})g_{rs}^{(nml)}\left(\frac{Z_{rs}-1-(s-1)\cdot3^{Kdn}}{3^{Kdn}}\right)+g_{\min}\\
&=(g_{\max}-g_{\min})\sum_{i=1}^{\infty}2^{-i}\theta_{rsi}^{(Z_{rs})}+g_{\min},
\end{align*}
we can now derive that
\begin{align*}
&|\widetilde{f}_{rs}(\boldsymbol{X})-{f}_{rs}(\boldsymbol{X})|\\
&=\left|(g_{\max}-g_{\min})\sum_{i=1}^{H}2^{-i}\theta_{rsi}^{(Z_{rs})}+g_{\min}-\left((g_{\max}-g_{\min})\sum_{i=1}^{\infty}2^{-i}\theta_{rsi}^{(Z_{rs})}+g_{\min}\right)\right.\\
&\left.\quad+g_{rs}\left(3\sum_{q=1}^n\sum_{p=1}^{d}\frac{1}{3^{(p-1)n+q}}\phi_K(x_{pq})\right)-g_{rs}\left(3\sum_{q=1}^n\sum_{p=1}^{d}\frac{1}{3^{(p-1)n+q}}\phi(x_{pq})\right)\right|\\
&\leq(g_{\max}-g_{\min})\left|\sum_{i=1}^{H}2^{-i}\theta_{rsi}^{(Z_{rs})}-\sum_{i=1}^{\infty}2^{-i}\theta_{rsi}^{(Z_{rs})}\right|\\
&\quad+\left|g_{rs}\left(3\sum_{q=1}^n\sum_{p=1}^{d}\frac{1}{3^{(p-1)n+q}}\phi_K(x_{pq})\right)-g_{rs}\left(3\sum_{q=1}^n\sum_{p=1}^{d}\frac{1}{3^{(p-1)n+q}}\phi(x_{pq})\right)\right|\\
&\leq(g_{\max}-g_{\min})\sum_{i=H+1}^{\infty}2^{-i}\theta_{rsi}^{(Z_{rs})}+2^{\beta}Q\left(3\sum_{q=1}^n\sum_{p=1}^{d}\frac{1}{3^{(p-1)n+q}}|\phi_K(x_{pq})-\phi(x_{pq})|\right)^{\log_32\cdot\frac{\beta}{dn}}\\
&\leq\frac{g_{\max}-g_{\min}}{2^H}+\frac{2^{\beta}Q}{2^{(K+2)\beta}},
\end{align*}
where in the third step we the fact $g_{rs}\in\mathcal{H}_{2^\beta Q}^{\frac{\beta \log 2}{d \log 3}}(\mathcal{C})$ from Proposition \ref{KST} and in the fourth step we use the following estimate:
\begin{align*}
|\phi_K(x)-\phi(x)|=2\sum_{j=K+1}^{\infty}a_{j}^x3^{-dn(j-1)-1}\leq\frac{2}{3^{2dn+1}-3^{dn+1}}\frac{1}{3^{dnK}}.
\end{align*}

\section{Proof of Lemma \ref{mmr-iw}}
We only present a proof of (1). The proof of (2) is exactly the same. Let
\begin{align*}
\Lambda:=\left\{3^{Kdn+1}\sum_{q=1}^n\sum_{p=1}^{d}\frac{1}{3^{(p-1)n+q}}\phi_K(x_{pq})+1+(s-1)\cdot3^{Kdn}:\boldsymbol{x}\in[0,1]^{d\times n},s\in[n]\right\}.
\end{align*}
Given $r\in[d]$, by Lemma \ref{NP-app} (for the case of (2), we apply Lemma \ref{RC-app}), there exist $w_r^{(0)},w_r^{(1)},w_r^{(2)},w_r^{(3)},w_r^{(4)}$ such that for each 
$$m=3^{Kdn+1}\sum_{q=1}^n\sum_{p=1}^{d}\frac{1}{3^{(p-1)n+q}}\phi_K(x_{pq})+1+(s-1)\cdot3^{Kdn}\in\Lambda,$$
there holds
\begin{align*}
\left|w_r^{(3)}\sigma_P\left(w_r^{(2)}\sigma_{NP}\left(w_r^{(0)}+w_r^{(1)}m\right)\right)+w_r^{(4)}-g_{rs}\left(\frac{m-1-(s-1)\cdot3^{Kdn}}{3^{Kdn}}\right)\right|\leq\delta.
\end{align*}
Define the memorization function 
\begin{align*}
f_{FF,r}^{(mmr)}(x):=w_r^{(3)}\sigma_P\left(w_r^{(2)}\sigma_{NP}\left(w_r^{(0)}+w_r^{(1)}x\right)\right)+w_r^{(4)},\quad r\in[d],
\end{align*}
then we have
\begin{align*}
\left|f_{FF,r}^{(mmr)}(m)-g_{rs}\left(\frac{m-1-(s-1)\cdot3^{Kdn}}{3^{Kdn}}\right)\right|\leq\delta.
\end{align*}
$\boldsymbol{f}_{FF}^{(mmr)}(\boldsymbol{x})$ is formed by arranging $\{f_{FF,r}^{(mmr)}\}_{r\in[d]}$ in parallel:
\begin{align*}
\boldsymbol{f}_{FF}^{(mmr)}(\boldsymbol{x}):=
\begin{pmatrix}
f_{FF,1}^{(mmr)}(x_1) & f_{FF,2}^{(mmr)}(x_2) & \cdots & f_{FF,d}^{(mmr)}(x_d)
\end{pmatrix}^T.
\end{align*}
If follows that
\begin{align*}
\boldsymbol{\widetilde{f}}_{:,s}&=\boldsymbol{f}_{FF}^{(mmr)}(\boldsymbol{Z}_{:,s})=
\begin{pmatrix}
f_{FF,1}^{(mmr)}(Z_{1s}) \\
f_{FF,2}^{(mmr)}(Z_{2s}) \\
\vdots \\
f_{FF,d}^{(mmr)}(Z_{ds})
\end{pmatrix}
=\begin{pmatrix}
w_1^{(3)}\sigma_P\left(w_1^{(2)}\sigma_{NP}\left(w_1^{(0)}+w_1^{(1)}Z_{1s}\right)\right)+w_1^{(4)} \\
w_2^{(3)}\sigma_P\left(w_2^{(2)}\sigma_{NP}\left(w_2^{(0)}+w_2^{(1)}Z_{1s}\right)\right)+w_2^{(4)} \\
\vdots \\
w_d^{(3)}\sigma_P\left(w_d^{(2)}\sigma_{NP}\left(w_d^{(0)}+w_d^{(1)}Z_{1s}\right)\right)+w_d^{(4)}
\end{pmatrix}.
\end{align*}
Recall that
\begin{align*}
f_{rs}\left(\boldsymbol{X}\right)=g_{rs}\left(3\sum_{q=1}^n\sum_{p=1}^{d}\frac{1}{3^{(p-1)n+q}}\phi(x_{pq})\right)
\end{align*}
and make use of the following equality:
\begin{align*}
g_{rs}\left(3\sum_{q=1}^n\sum_{p=1}^{d}\frac{1}{3^{(p-1)n+q}}\phi_K(x_{pq})\right)
=g_{rs}\left(\frac{Z_{rs}-1-(s-1)\cdot3^{Kdn}}{3^{Kdn}}\right),
\end{align*}
we can now derive that
\begin{align*}
&|\widetilde{f}_{rs}(\boldsymbol{X})-{f}_{rs}(\boldsymbol{X})|\\
&=\left|w_r^{(3)}\sigma_P\left(w_r^{(2)}\sigma_{NP}\left(w_r^{(0)}+w_r^{(1)}Z_{rs}\right)\right)+w_r^{(4)}-g_{rs}\left(\frac{{Z}_{rs}-1-(s-1)\cdot3^{Kdn}}{3^{Kdn}}\right)\right.\\
&\left.\quad\ +g_{rs}\left(3\sum_{q=1}^n\sum_{p=1}^{d}\frac{1}{3^{(p-1)n+q}}\phi_K(x_{pq})\right)-g_{rs}\left(3\sum_{q=1}^n\sum_{p=1}^{d}\frac{1}{3^{(p-1)n+q}}\phi(x_{pq})\right)\right|\\
&\leq\left|w_r^{(3)}\sigma_P\left(w_r^{(2)}\sigma_{NP}\left(w_r^{(0)}+w_r^{(1)}Z_{1s}\right)\right)+w_r^{(4)}-g_{rs}\left(\frac{{Z}_{rs}-1-(s-1)\cdot3^{Kdn}}{3^{Kdn}}\right)\right|\\
&\quad\ +\left|g_{rs}\left(3\sum_{q=1}^n\sum_{p=1}^{d}\frac{1}{3^{(p-1)n+q}}\phi_K(x_{pq})\right)-g_{rs}\left(3\sum_{q=1}^n\sum_{p=1}^{d}\frac{1}{3^{(p-1)n+q}}\phi(x_{pq})\right)\right|\\
&\leq\delta+2^{\beta}Q\left(3\sum_{q=1}^n\sum_{p=1}^{d}\frac{1}{3^{(p-1)n+q}}|\phi_K(x_{pq})-\phi(x_{pq})|\right)^{\log_32\cdot\frac{\beta}{dn}}\\
&\leq\delta+\frac{2^{\beta}Q}{2^{(K+2)\beta}},
\end{align*}
where in the third step we the fact $g_{rs}\in\mathcal{H}_{2^\beta Q}^{\frac{\beta \log 2}{d \log 3}}(\mathcal{C})$ from Proposition \ref{KST} and in the fourth step we use the following estimate:
\begin{align*}
|\phi_K(x)-\phi(x)|=2\sum_{j=K+1}^{\infty}a_{j}^x3^{-dn(j-1)-1}\leq\frac{2}{3^{2dn+1}-3^{dn+1}}\frac{1}{3^{dnK}}.
\end{align*}

\section{Implementation of $\phi_K$}

The following lemma suggests the existence of a ReLU-feedforward neural network function $f_{FF}^{(R-pi)}$ that can realize a piecewise function generated by $\phi_K$ at almost all $x\in[0,2n-1]$.

\begin{lemma}\label{Rpi}
Let $p\in[1,\infty)$. There exists a ReLU-feedforward neural network function $f_{FF}^{(R-pi)}:\mathbb{R}\to\mathbb{R}$ with depth $2K$ and width $4n$ such that for $x\in[2q,2q+1]\backslash(\Omega^{(flow)}+2q),q\in\{0,1,\dots,n-1\}$, there holds
\begin{align*}
f_{FF}^{(R-pi)}(x)=\frac{1}{3^{q-1}}\phi_K(x-2q)+\frac{9}{2}-\frac{1}{2}\frac{1}{3^{q-2}}.
\end{align*}
Here $\Omega^{(flaw)}\subset[0,1]$ is some region with the Lebesgue measure not greater than $2^{-K\beta p}$.
\end{lemma}
\begin{proof}
It is shown in Theorem 3 of \cite{schmidt2021kolmogorov} that there exists a ReLU-feedforward neural network function $f_{FF}^{(inner)}$ with depth $2K$ and width $4$ such that 
\begin{align*}
f_{FF}^{(inner)}(x)=\left\{\begin{matrix}
0, & x\in(-\infty,0);\\
\phi_K(x), & x\in[0,1]\backslash\Omega^{(flaw)};\\
1, & x\in(1,\infty),
\end{matrix}\right.
\end{align*}
where $\Omega^{(flaw)}\subset[0,1]$ is some region with the Lebesgue measure not greater than $2^{-k\beta p}$. Then the piecewise inner function $f_{FF}^{(R-pi)}$ can be constructed in the following way:
\begin{align*}
f_{FF}^{(R-pi)}(x):&=\begin{pmatrix}
3  & 1 & \cdots & \frac{1}{3^{n-2}}
\end{pmatrix}\begin{pmatrix}
f_{FF}^{(inner)}\left(x\right) \\
f_{FF}^{(inner)}\left(x-2\right) \\
\vdots \\
f_{FF}^{(inner)}\left(x-2(n-1)\right)
\end{pmatrix}=\sum_{q=0}^{n-1}\frac{1}{3^{q-1}}f_{FF}^{(inner)}\left(x-2q\right).
\end{align*}
It is easy to verify that for $x\in[2q,2q+1]\backslash(\Omega^{(flow)}+2q),q\in\{0,1,\dots,n-1\}$, there holds
\begin{align*}
f_{FF}^{(R-pi)}(x)=\frac{1}{3^{q-1}}\phi_K(x-2q)+\frac{9}{2}-\frac{1}{2}\frac{1}{3^{q-2}}.
\end{align*}
\end{proof}

By incorporating the floor function as an activation function in the feedforward neural network, we can extend the above result to hold for all $x \in [0,2n-1]$, and the required depth and width of the neural network function are reduced.

\begin{lemma}\label{FRpi}
There exists a floor-ReLU-feedforward neural network function $f_{FF}^{(FR-pi)}:\mathbb{R}\to\mathbb{R}$ with depth $K+3$ and width $3n$
such that for $x\in[2q,2q+1],q\in\{0,1,\dots,n-1\}$, there holds
\begin{align*}
f_{FF}^{(FR-pi)}(x)=
\frac{1}{3^{q-1}}\phi_K(x-2q). 
\end{align*}
\end{lemma}
\begin{proof}
The proof is divided into two steps. In the first step, we show that $\phi_K(x)$ can be implemented by a floor-feedforward network function $f_{FF}^{(inner)}$(we call it as inner function) with depth $K+1$ and width $3$ on $[0,1]$ without any error. The second step is to adopt $f_{FF}^{(inner)}$ to construct the piecewise inner function $f_{FF}^{(FR-pi)}$.

Given $x=\left[0.a_1^x a_2^x \ldots\right]_2\in[0,1]$, define $\{T_j\}_{j=1}^K$ to be
\begin{align}\label{FRpi1}
T_1&=2x;\nonumber\\
T_j&=2\left[\frac{1}{2^{K-j+1}}\sigma_F(2^{K-j+1}T_{j-1})-\sigma_F(T_{j-1})\right],\quad j\in\{2,3,\dots,K\}.
\end{align}
It can be verified by the induction that, on the right-hand side in the definition of $T_j$ for $j\geq2$, the first term is actually $T_{j-1}$ and the second term is the integer part of $T_{j-1}$ and hence
\begin{align*}
T_j=[a_j^x.a_{j+1}^x\dots a_K^x]_2,\quad j\in\{2,3,\dots,K\}.
\end{align*}
We then define another sequence $\{S_j\}_{j=1}^{K}$:
\begin{align}\label{FRpi2}
S_0&=0;\nonumber\\
S_{j}&=2\cdot3^{dn(K-j)}\sigma_F(T_{j})+\sigma_F(S_{j-1}),\quad j\in\{1,2,\cdots,K\}.
\end{align}
It is not hard to find that 
\begin{align*}
S_{j}&=2\sum_{j'=1}^ja_{j'}^x3^{dn(K-j')}.
\end{align*}
Let
\begin{align*}
f_0&=x;\\
f_j&=
\begin{pmatrix}
\sigma_F(2^{K-j}T_j) \\
\sigma_F(T_j) \\
\sigma_F(S_{j-1})
\end{pmatrix},\quad j\in\{1,2,\dots,K\}.
\end{align*}
By (\ref{FRpi1})(\ref{FRpi2}) we conclude that
\begin{align}\label{FRpi3}
f_0&=x;\nonumber\\
f_{j}&=\sigma_F(A_{j}f_{j-1}),\quad j\in\{1,2,\dots,K\},
\end{align}
where
\begin{align*}
A_1=\begin{pmatrix}
2^K \\
2 \\
0
\end{pmatrix};\quad
A_j=\begin{pmatrix}
1  & -2^{K-j+1} & 0\\
\frac{1}{2^{K-j}}  & -2 & 0\\
0  & 2\cdot3^{dn(K-j+1)} & 1
\end{pmatrix},j\in\{2,3,\dots,K\}.
\end{align*}
Define
\begin{align}\label{FRpi4}
f_{FF}^{(inner)}&=A_{K+1}f_K
\end{align}
with
\begin{align*}
A_{K+1}=\frac{1}{3^{dn(K-1)+1}}\begin{pmatrix}
0  & 2 & 1
\end{pmatrix}.
\end{align*}
By (\ref{FRpi3})(\ref{FRpi4}) we know that $f_{FF}^{(inner)}$ is a floor-feedforward network function with depth $K+1$ and width $3$ and for any $x\in[0,1]$,
\begin{align*}
f_{FF}^{(inner)}&=\frac{1}{3^{dn(K-1)+1}}[2\sigma_F(T_{K})+\sigma_F(S_{K-1})]=\frac{1}{3^{dn(K-1)+1}}\left[2a_K^K+2\sum_{j=1}^{K-1}a_{j}^x3^{dn(K-j)}\right]\\
&=\frac{1}{3^{dn(K-1)+1}}2\sum_{j=1}^Ka_{j}^x3^{dn(K-j)}=\phi_K(x).
\end{align*}
Figure \ref{D2} illustrates the flowchart for computing \(\phi_K(x)\) when \(K = 4\).

\begin{figure}
\vskip 0.2in
\begin{center}
\begin{tikzpicture}[>=stealth,
  node distance=1.5cm,         
  every node/.style={font=\normalsize}]

  \node (x)           at (0,0)   {$x$};
  \node (T1)          at (1.5,0) {$T_1$};
  \node (T2)          at (3,0)   {$T_2$};
  \node (T3)          at (4.5,0) {$T_3$};
  \node (T4)          at (6,0)   {$T_4$};

  \node (S1) at (1.5,1.5) {$S_1$};
  \node (S2) at (3,1.5)   {$S_2$};
  \node (S3) at (4.5,1.5) {$S_3$};
  \node (S4) at (6,1.5)   {$S_4$};

  \node (phi) at (9,1.5)   {$\phi_4(x)$};

  \draw[->] (x)  -- (T1);
  \draw[->] (T1) -- (T2);
  \draw[->] (T2) -- (T3);
  \draw[->] (T3) -- (T4);

  \draw[->] (T1) -- (S1);
  \draw[->] (T2) -- (S2);
  \draw[->] (T3) -- (S3);
  \draw[->] (T4) -- (S4);

  \draw[->] (S1) -- (S2);
  \draw[->] (S2) -- (S3);
  \draw[->] (S3) -- (S4);

  \draw[->] (S4) -- node[above]
    {$\times\dfrac{1}{3^{3dn+1}}$} (phi);

\end{tikzpicture}
\caption{The flowchart for computing \(\phi_K(x)\) when \(K = 4\).}
\label{D2}
\end{center}
\vskip -0.2in
\end{figure}
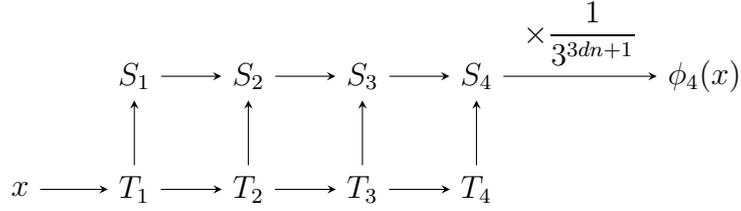

Now we adpot $f_{FF}^{(inner)}$ to construct $f_{FF}^{(FR-pi)}$. Firstly, we use a ReLU-feedforward network to filter out $(-\infty,0)$ and $(1,\infty)$, the intervals that we are not concerned with:
\begin{align*}
f_{FF}^{(filter)}(x):=-\sigma_R(-\sigma_R(x)+1)+1=\left\{\begin{matrix}
0,  & x\in(-\infty,0);\\
x,  & x\in[0,1];\\
1,  & x\in(1,\infty).
\end{matrix}\right.
\end{align*}
The composition of $f_{FF}^{(inner)}$ and $f_{FF}^{(filter)}$ yields
\begin{align*}
f_{FF}^{(inner)}\left(f_{FF}^{(filter)}(x)\right)=\left\{\begin{matrix}
\phi_K(x),  & x\in[0,1]; \\
0,  & x\notin[0,1].
\end{matrix}\right.
\end{align*}
We can now construct the piecewise inner function $f_{FF}^{(FR-pi)}$ in the following way:
\begin{align*}
f_{FF}^{(FR-pi)}(x):&=\begin{pmatrix}
3  & 1 & \cdots & \frac{1}{3^{n-2}}
\end{pmatrix}\begin{pmatrix}
f_{FF}^{(inner)}\left(f_{FF}^{(filter)}(x)\right) \\
f_{FF}^{(inner)}\left(f_{FF}^{(filter)}(x-2)\right) \\
\vdots \\
f_{FF}^{(inner)}\left(f_{FF}^{(filter)}(x-2(n-1))\right)
\end{pmatrix}\\
&=\sum_{q=0}^{n-1}\frac{1}{3^{q-1}}f_{FF}^{(inner)}\left(f_{FF}^{(filter)}(x-2q)\right).
\end{align*}
It is easy to verify that for $x\in[2q,2q+1],q\in\{0,1,\dots,n-1\}$, there holds
\begin{align*}
f_{FF}^{(FR-pi)}(x)=
\frac{1}{3^{q-1}}\phi_K(x-2q). 
\end{align*}

\end{proof}

\section{Memorization of Feedforward Neural Networks}

In this section, we present some memorization results of feedforward neural networks. These results demonstrate the remarkable memorization capacity of feedforward neural networks: a very small neural network can memorize a vast dataset. To some extent, these findings theoretically reveal the reasons behind the tremendous success of neural networks in practical applications.

The memorization issue we are referring to here can be exact memorization or approximate memorization. That is, for a given set of data pairs $\{(x_m, y_m)\}_{m=1}^M$, the memory capability of feedforward neural networks implies the existence of a feedforward neural network function $f_{FF}:\mathbb{R}\to\mathbb{R}$ such that 
\begin{align*}
f_{FF}(x_m) = y_m,\quad m\in[M], 
\end{align*}
where "$=$" can be replaced by "$\approx$".


We first provide some results of exact binary memorization tasks.
\begin{lemma}\label{binary-app}
Let $M\in\mathbb{N}_{\geq1}$. For any $\{\theta_m\}_{m\in[M]}\subset\{0,1\}^M$, there exists a feedforward neural network function $f_{FF}^{(mmr)}:\mathbb{R}\to\mathbb{R}$ such that
\begin{align*}
f_{FF}^{(mmr)}(m)=\theta_m,\quad\forall m\in[M].
\end{align*}
$f_{FF}^{(mmr)}$ can be in one of the following feedforward neural network classes:
\begin{enumerate}[label=(\arabic*)]
\item $\mathcal{FF}\left(7L-2,2\left\lceil M^{1/L}\right\rceil+2,\{\text{ReLU, floor}\}\right)$, where $L$ can be any positive integer number;
\item $\mathcal{FF}(6,3,\{\text{ReLU, floor}, 2^x\})$;
\item $\mathcal{FF}(4,2,\{\text{ReLU, sine}, 2^x\})$;
\item $\mathcal{FF}(5,1,\{\text{ReLU, cosine}, 3^x\})$.
\end{enumerate}
\end{lemma}
\begin{proof}
(1) See Proposition 1 in \cite{shen2021deep}.

(2) See Proposition 3.2 in \cite{shen2021neural}. Note that $\sigma_3$ there is defined to be $\mathcal{T}(x-\lfloor x\rfloor-\frac{1}{2})$ with $\mathcal{T}(x)$ defined as $\mathbbm{1}_{x\geq0}$. Note that
\begin{align*}
\mathbbm{1}_{x\geq0}&=\sigma_F(-\sigma_R(-\sigma_R(x+1)+1)+1),\\
x-\lfloor x\rfloor-\frac{1}{2}&=\sigma_R(x)+\sigma_R(-x)-\sigma_F(x)-\frac{1}{2},
\end{align*}
therefore $\sigma_3$ in Proposition 3.2 in \cite{shen2021neural} can be implemented by a ReLU-floor feedforward network function with depth $5$ and width $3$.

(3) See Lemma 3.1 in \cite{jiao2023deep}.

(4) See Proposition 4.2 in \cite{shen2021neural}. Note that $\rho_3$ there is defined to be $\mathcal{\widetilde{T}}(\cos(2\pi x))$ with $\widetilde{T}(x)$ defined as a piecewise linear function:
\begin{align*}
\widetilde{\mathcal{T}}(x)=\left\{\begin{matrix}
0,  & x\in(\cos(\frac{4\pi}{9}),\infty);\\
1-x/\cos(\frac{4\pi}{9}),  & x\in[0,\cos(\frac{4\pi}{9})];\\
1,  & x\in(-\infty,0).
\end{matrix}\right.
\end{align*}
This piecewise linear function can be furthered reproduced by a ReLU feedforward network function with depth $3$ and width $1$:
\begin{align*}
\widetilde{\mathcal{T}}(x)=-\sigma_R\left(-\sigma_R\left(-\frac{x}{\cos\left(\frac{4\pi}{9}\right)}+1\right)+1\right)+1.
\end{align*}
\end{proof}

Next, we introduce some results of memorization tasks where the labels can take on arbitrary values. These results make use of a well-known fact that an irrational
winding on the torus is dense. Before we state this fact, we need to know the concept of rationally independent.

\begin{definition}
For any $\{a_m\}_{m\in[M]}\subset \mathbb{R}$, they are called rationally independent if they are linearly independent over the rational numbers $\mathbb{Q}$. That is, if there exist $\lambda_1, \lambda_2, \cdots, \lambda_M \in \mathbb{Q}$ such that $\sum_{m=1}^M\lambda_m  a_m=0$, then $\lambda_1=\lambda_2=\cdots=\lambda_M=0$. 
\end{definition}

The following two lemmas provide two methods for generating rationally independent numbers.
\begin{lemma}[\cite{yarotsky2021elementary}, Lemma 1]\label{NP-ri}
Let $M\in\mathbb{N}_{\geq1}$. There exists $w^{(1)},w^{(0)}\in\mathbb{R}$ such that $\left\{\sigma_{NP}\left(w^{(1)} m+w^{(0)}\right)\right\}_{m\in[M]}$ are rationally independent. 
\end{lemma}

\begin{lemma}[\cite{zhang2022deep}, Lemma 18]\label{RC-ri}
Let $M\in\mathbb{N}_{\geq1}$. $\left\{\sigma_{RC}(m)\right\}_{m\in[M]}
$
are rationally independent.
\end{lemma}

\begin{remark}
Lemma \ref{RC-ri} is a simplified version of Lemma 18 in \cite{zhang2022deep}, as the latter allows for replacing the sequence of positive integers $\{m\}_{m\in[M]}$ with any sequence of distinct rational numbers $\{r_m\}_{m\in[M]}$.
\end{remark}

The following proposition describes that an irrational
winding on the torus is dense. For a proof, the readers are referred to Lemma 2 in \cite{yarotsky2021elementary} and Lemma 19 in \cite{zhang2022deep}.
\begin{proposition}\label{irrational
winding}
Let $M\in\mathbb{N}_{\geq1}$.
Given any rationally independent numbers $\{a_m\}_{m\in[M]}$ and an arbitrary periodic function $\sigma_P: \mathbb{R} \rightarrow \mathbb{R}$ with period $T$, i.e., $\sigma_P(x+T)=\sigma_P(x)$ for any $x \in \mathbb{R}$, assume there exist $x_1, x_2 \in \mathbb{R}$ with $0<x_2-x_1<T$ such that $g$ is continuous on $\left[x_1, x_2\right]$. Then the following set
$$
\left\{\left(\sigma_P\left(w a_1\right), \sigma_P\left(w a_2\right), \cdots, \sigma_P\left(w a_M\right)\right)^T: w \in \mathbb{R}\right\}
$$
is dense in $\left[\sigma_P^{(\min)}, \sigma_P^{(\max)}\right]^M$, where $\sigma_P^{(\min)}:=\min _{x \in\left[x_1, x_2\right]} \sigma_P(x)$ and $\sigma_P^{(\max)}:=\max _{x \in\left[x_1, x_2\right]} \sigma_P(x)$.
In the case of $\sigma_P^{(\min)}<\sigma_P^{(\max)}$, the following set
$$
\left\{\left(u \cdot \sigma_P\left(w a_1\right)+v, u \cdot \sigma_P\left(w a_2\right)+v, \cdots, u \cdot \sigma_P\left(w a_M\right)+v\right)^T: u, v, w \in \mathbb{R}\right\}
$$
is dense in $\mathbb{R}^M$.
\end{proposition}

With these tools, we are able to construct feedforward neural network that memorize the sequence of positive integers $\{m\}_{m\in[M]}$ to given real-valued labels. We can see that the networks in the following lemmas are fixed size and they can achieve any accuracy when memorizing integers.
\begin{lemma}\label{NP-app}
Let $M\in\mathbb{N}_{\geq1}$. For any $\epsilon>0$ and any $\boldsymbol{\xi}\in\mathbb{R}^M$, there exist $w_r^{(0)},w_r^{(1)},w_r^{(2)},w_r^{(3)},w_r^{(4)}\in\mathbb{R}$ such that for $m\in[M]$, there holds
\begin{align*}
\left|w_r^{(3)}\sigma_P\left(w_r^{(2)}\sigma_{NP}\left(w_r^{(0)}+w_r^{(1)}m\right)\right)+w_r^{(4)}-\xi_m\right|\leq\epsilon.
\end{align*}
\end{lemma}
\begin{proof}
A direct result from Lemma \ref{NP-ri} and Proposition \ref{irrational
winding}.
\end{proof}

\begin{lemma}\label{RC-app}
Let $M\in\mathbb{N}_{\geq1}$. For any $\epsilon>0$ and any $\boldsymbol{\xi}\in\mathbb{R}^M$, there exist $w^{(0)}, w^{(1)}, w^{(2)} \in \mathbb{R}$ such that for $m\in[M]$, there holds
\begin{align*}
\left|w_{r}^{(1)}\sigma_{P}\left(w_{r}^{(0)}\sigma_{RC}\left(m\right)\right)+w_{r}^{(2)}-\xi_m\right|\leq\epsilon.
\end{align*}
\end{lemma}
\begin{proof}
A direct result from Lemma \ref{RC-ri} and Proposition \ref{irrational
winding}.
\end{proof}

\section{Comparison of Related Work}

In the table below, we have listed our results along with a comparison to some related work.

\begin{table}[ht]
\caption{Comparison of existing and our results on the approximation capabilities of different Transformer structures, where $\varepsilon \in (0,1)$ denotes the approximation error.}\label{comp}
\centering
\resizebox{\textwidth}{!}{
\begin{tabular}{p{2.3cm}<{\centering\arraybackslash} p{1.35cm}<{\centering\arraybackslash} p{1.6cm}<{\centering\arraybackslash} p{1.0cm}<{\centering\arraybackslash} p{2.3cm}<{\centering\arraybackslash} p{2.45cm}<{\centering\arraybackslash} p{2.25cm}<{\centering\arraybackslash} p{2.15cm}<{\centering\arraybackslash}}
\toprule[1.5pt]
& \textbf{Type} & \textbf{Target Function}$^1$ & \textbf{Metric} & \textbf{Activations in Self-Attention Layers}$^2$ & \textbf{Activations in Feedforward Layers} & \textbf{Width}$^3$ & \textbf{Depth} \\
\midrule[1pt]
\citet{DBLP:conf/iclr/YunBRRK20} &  Universality & $\mathcal{C}^0$ & $L^p$ & $\sigma_{S}$ with bias & $\sigma_{R}$ & \multirow{3}{*}{
\begin{tikzpicture}[baseline=(current bounding box.center)]
    \draw (0,3.8) -- (4.9,0);
\end{tikzpicture}
} & \\
\cmidrule{1-6}
\citet{kajitsukatransformers} & Universality & $\mathcal{C}^0$ & $L^p$ & $\sigma_{S}$ & $\sigma_{R}$ &  &  \\
\cmidrule{1-6}
\citet{fang2022attention} & Universality & $\mathcal{C}^0$ & $L^\infty$ & $\sigma_{H}$ & $\sigma_{R}$ &  &  \\
\midrule
\citet{gurevych2022rate} & Rate & $\mathcal{F}(\mathcal{P})$ & $L^\infty$ & $X \odot \sigma_H(X)$ & $\sigma_{R}$ & $\mathcal{O}(\max\limits_{(p, K)\in \mathcal{P}} \varepsilon^{-K/p})$ & $\mathcal{O}(1)$ \\
\midrule
\vspace{-14pt}\citet{jiao2024convergence} & Rate & \makecell[c]{$\mathcal{H}^\beta$\\ $\mathcal{C}^m$} & $L^\infty$ & $X \odot \sigma_H(X)$ & $\sigma_{R}$ & \makecell[c]{$\mathcal{O}(\varepsilon^{-dn/\beta})$\\ $\mathcal{O}(\varepsilon^{-dn/m})$} & 
\makecell[c]{$\mathcal{O}(\log\frac{1}{\varepsilon} )$\\ $\mathcal{O}(\log \frac{1}{\varepsilon})$} \\
\midrule
\citet{takakura2023approximation} & Rate & $\mathcal{B}_{p, \theta}^{\gamma(\cdot; a)}$ & $L^2$ & $\sigma_{S}$ & $\sigma_{R}$ & $\mathcal{O}(\varepsilon^{-1/a^\dagger} \log \frac{1}{\varepsilon})$ & $\mathcal{O}((\log \frac{1}{\varepsilon})^2)$ \\
\midrule
\citet{havrilla2024understanding} & Rate & $\mathcal{H}^\beta$ & $L^\infty$ & $\sigma_{R}$ & $\sigma_{R}$ & {$\mathcal{O}(\varepsilon^{-dn/\beta})$} & $\mathcal{O}(\log \frac{1}{\varepsilon})$ \\
\midrule
\citet{jiaolaisunwangyan} & Rate & $\mathcal{H}^\beta$ & $L^\infty$ & $\sigma_{S}$ & $\sigma_{R}$ & {$\mathcal{O}(\varepsilon^{-dn/\beta})$} & $\mathcal{O}(\log \frac{1}{\varepsilon})$ \\
\midrule[1pt]
\makecell[c]{\textbf{Ours}\\ (Theorem \ref{Linfty})} & Rate & $\mathcal{H}^\beta$ & $L^\infty$ & $\sigma_{S}$ & \makecell[c]{$\sigma_{R}, \lfloor\cdot\rfloor$\\ $\sigma_{R}, \lfloor\cdot\rfloor, 2^x$\\ $\sigma_{R}, \lfloor\cdot\rfloor, \sigma_{NP}$\\ $\sigma_{R}, \lfloor\cdot\rfloor, \sigma_{RC}$} & \makecell[c]{\small$\mathcal{O}(\varepsilon^{-\frac{2}{\beta}} \log \frac{1}{\varepsilon})$\\ $\mathcal{O}(\log \frac{1}{\varepsilon})$\\ $\mathcal{O}(1)$\\ $\mathcal{O}(1)$} & \makecell[c]{$\mathcal{O}(\log \frac{1}{\varepsilon})$\\ $\mathcal{O}(\log \frac{1}{\varepsilon})$\\ $\mathcal{O}(\log \frac{1}{\varepsilon})$\\ $\mathcal{O}(\log \frac{1}{\varepsilon})$} \\
\midrule
\makecell[c]{\textbf{Ours}\\ (Theorem \ref{Lp})} & Rate & $\mathcal{H}^\beta$ & $L^p$ & $\sigma_{S}$ & \makecell[c]{$\sigma_{R}, sine, 2^x$\\ $\sigma_{R}, cosine, 3^x$\\ $\sigma_{R}, \sigma_{P}, \sigma_{NP}$\\ $\sigma_{R}, \sigma_{P}, \sigma_{RC}$} & \makecell[c]{$\mathcal{O}(\log \frac{1}{\varepsilon})$\\ $\mathcal{O}(\log \frac{1}{\varepsilon})$\\ $\mathcal{O}(1)$\\ $\mathcal{O}(1)$} & \makecell[c]{$\mathcal{O}(\log \frac{1}{\varepsilon})$\\ $\mathcal{O}(\log \frac{1}{\varepsilon})$\\ $\mathcal{O}(\log \frac{1}{\varepsilon})$\\ $\mathcal{O}(\log \frac{1}{\varepsilon})$} \\
\bottomrule[1.5pt]
\end{tabular}
}
\vspace{0.2em} \raggedright
1. $\mathcal{H}^\beta$ denotes the space of $\beta$-Hölder continuous functions with compact support and bounded norm. The space $\mathcal{C}^m$ consists of all functions whose first $m$ derivatives exist and are continuous, and $\mathcal{C}^0$ denotes the space of continuous functions. $\mathcal{B}_{p, \theta}^{\gamma(\cdot; a)}$ refers to the Besov space with mixed and anisotropic smoothness defined in \citet{takakura2023approximation}. $\mathcal{F}(\mathcal{P})$ denotes the hierarchical composition model with smoothness and order constraint $\mathcal{P}$ defined in \citet{gurevych2022rate}.
2. We denote the softmax activation function by $\sigma_{S}$, the hardmax activation function by $\sigma_{H}$, and the ReLU activation function, which always operates component-wise, by $\sigma_{R}$. The symbol $\odot$ represents the Hadamard product.
3. Following \citet{kim2023provable}, the width of the Transformer network is defined as $\max\{hs, W\}$.
\end{table}

\end{document}